\documentclass{ws-ijns}
\usepackage{graphicx}
\usepackage{url}
\usepackage[english]{babel}
\usepackage[super,sort,compress]{cite}

\usepackage[caption=false]{subfig}
\usepackage{enumitem}
\usepackage{amsmath,amssymb}
\usepackage{amsfonts}       %
\usepackage{nicefrac}       % compact symbols for 1/2, etc.
\usepackage{pgfplots}
\pgfplotsset{compat=1.16}
\setlist{leftmargin=\parindent}
\usepackage{multirow}
\newcommand{\CH}[1]{}

\newcommand{\IN}{\mathbb{N}}

\newcommand{\IR}{\mathbb{R}}
\newcommand{\bs}[1]{\boldsymbol{#1}}

\begin{document}

\catchline{0}{0}{2021}{}{}

\markboth{Kov\'acs et al.}{VPNet: Variable Projection Networks}

\title{VPNET: VARIABLE PROJECTION NETWORKS}
%\LaTeX\footnote{For the title, try not to use more than}

\author{P\'ETER KOV\'ACS\footnote{Corresponding author, e-mail: \protect\url{kovika@inf.elte.hu}, website: \protect\url{http://numanal.inf.elte.hu/~kovika}}}
\address{Department of Numerical Analysis, E\"otv\"os Lor\'and University, \\ P\'azm\'any P\'eter stny. 1/C, Budapest, 1117, Hungary}

\author{GERG\H{O} BOGN\'AR}
\address{JKU LIT SAL eSPML Lab, Institute of Signal Processing, Johannes Kepler University Linz,\\
Altenberger str. 69, Linz, 4040, Austria\\
JKU LIT SAL eSPML Lab, Silicon Austria Labs,\\
Altenberger str. 69, Linz, 4040, Austria\\
Department of Numerical Analysis, E\"otv\"os Lor\'and University, \\ P\'azm\'any P\'eter stny. 1/C, Budapest, 1117, Hungary}

\author{CHRISTIAN HUBER}
\address{Embedded AI research group, Silicon Austria Labs GmbH,\\
Altenberger str. 69, Linz, 4040, Austria}

\author{MARIO HUEMER}
\address{JKU LIT SAL eSPML Lab, Institute of Signal Processing, Johannes Kepler University Linz,\\
Altenberger str. 69, Linz, 4040, Austria\\
JKU LIT SAL eSPML Lab, Silicon Austria Labs,\\
Altenberger str. 69, Linz, 4040, Austria}

\maketitle

\begin{abstract}
We introduce VPNet, a novel model-driven neural network architecture based on variable projection (VP). Applying VP operators to neural networks results in learnable features, interpretable parameters, and compact network structures. This paper discusses the motivation and mathematical background of VPNet and presents experiments. The VPNet approach was evaluated in the context of signal processing, where we classified a synthetic dataset and real electrocardiogram (ECG) signals. Compared to fully connected and one-dimensional convolutional networks, VPNet offers fast learning ability and good accuracy at a low computational cost of both training and inference. Based on these advantages and the promising results obtained, we anticipate a profound impact on the broader field of signal processing, in particular on classification, regression and clustering problems.
\end{abstract}


\begin{multicols}{2}
\section{Introduction}
\label{sec:intro}
Until recently, signal processing was dominated by conventional model-based algorithms, which rely on mathematical and physical models of the real world.  
%In many cases these models simplify the original problem to a linear or to a nonlinear approximation, which is supported by a strong mathematical apparatus. 
They are inherently interpretable and often incorporate domain knowledge, such as statistical assumptions, smoothness, structure of the model space, and origin of the noise. However, this approach can become mathematically intractable if problems are complex. Machine learning (ML) provides an alternative approach to this challenge by building data-driven mathematical models. %This way, a better objective function value can be achieved, while the internal structure of the data is explored automatically by computer algorithms. 
Neural networks (NNs) and supervised learning in particular offer a proper framework for various signal-processing problems~\cite{deepreview2015}. Below, we briefly review a few recent trends that served as motivation for developing the proposed variable projection network (VPNet).

%Neural networks (NN) gives a popular and rapidly developing field, that has inherent impact on the evolution of ML itself \cite{deepreview2015}. NNs comprise concatenated multiple-input-single-output processors called artificial neurons, which are usually organized in layer-wise structures. NNs and supervised learning provide a proper framework for several signal processing problems. In the following, we briefly review a few recent trends that serve as motivations for VPNet.

A traditional ML approach is to decompose the problem into separate feature extraction and learning steps \cite{MLbook}. In this case, the data is preprocessed in order to extract static features based on the given domain knowledge. These features are inputs to conventional ML algorithms. Although the dimension of the original data is significantly reduced in the first step, these handcrafted features are usually suboptimal with respect to the whole learning process \cite{replearning}. Deep learning provides alternatives to the traditional approach, overcoming some of its drawbacks \cite{deepreview2015}.
\CH{start new}
Learned features of NNs can be used as input for non NN methods, like discriminant correlation filters, as well~\cite{yang_multi-object_2019}. Ref.~\refcite{diaz-vico_deep_2020} combined traditional kernel-based Support Vector Machines (SVMs)  with deep learning approaches.
Another common method would would be to use the features as input for one or multiple other NN for multi-target prediction~ \cite{reyes_performing_2019, mishra_neural_2020, girshick_fast_2015, ren_faster_2015}.
\CH{stop new}.

%In the past, traditional NN approaches used a single layer hidden in between the input and output layers. However, it turned out that the learning ability of NNs can be significantly improved by increasing the number of hidden layers \cite{science}, which laid down the foundations of deep neural networks (DNN). Theoretically, a DNN can be viewed as the composition of multi-level feature extraction and prediction layers, i.e. that DNNs have their own, built-in, learnable feature extraction method. Convolutional neural networks (CNN) are special, optionally deep architectures that are the leading ML methods in 2D and 3D image processing and computer vision. Here, the built-in feature extraction layers perform multiple convolutional filtering and dimension reduction (pooling) steps. A few recent examples are the AlexNet \cite{AlexNet}, ZFNet \cite{ZFNet}, GoogLeNet \cite{GoogLeNet}, and U-Net \cite{ronneberger2015u}. Besides CNNs, we mention the deep LSTM \cite{lstm}, AutoEncoder \cite{Goodfellow-et-al-2016}, and 1D CNN \cite{serkan2016} as well. 
Using more hidden layers in deep neural networks (DNNs) has increased the learning abilities of NNs \cite{science}. This enables DNNs to use the first layers for feature extraction and further layers for performing operations on the features learned. Convolutional neural networks (CNNs) are special, optionally deep architectures and are the leading ML approaches in 2D and 3D image processing and computer vision~\CH{added references}\cite{AlexNet, ZFNet, GoogLeNet, ronneberger2015u, manzanera_scaled_2019, leming_ensemble_2020}. Here, the built-in feature extraction layers perform multiple convolutional filtering and dimension-reduction (pooling) steps. 
%A few recent examples are the AlexNet \cite{AlexNet}, ZFNet \cite{ZFNet}, GoogLeNet \cite{GoogLeNet}, and U-Net \cite{ronneberger2015u}. Besides CNNs, we mention the deep LSTM \cite{lstm}, AutoEncoder \cite{Goodfellow-et-al-2016}, and 1D CNN \cite{serkan2016} as well.
Despite their advantages, DNNs and CNNs continue to raise several concerns. Their improved efficiency comes at the cost of higher computational complexity and numerical difficulties in the training process (see, e.g.,\ overfitting and divergence). Due to the large number of nonlinear connections between the model parameters, DNN and CNN approaches can be considered as black-box methods, where the parameters have no or little physical meaning and are difficult or impossible to interpret. 
%Actually, it is difficult to deduce the results achieved by DNNs, because the model parameters have no evident physical meaning, they have limited natural interpretation. 
Additionally, training these networks requires vast amounts of labeled data, which is problematic to collect in many applications, such as telecommunications \cite{DNNtelecom}, and biomedical engineering \cite{serkan2016, ECGsurvey}. Although data augmentation, transfer learning, outlier removal, and ensemble methods can mitigate this problem, reducing the data hunger of deep learning approaches is still a major challenge in this field.
%The numeric issues induce not only new network architectures, but also adaptive learning methods (see e.g. Adam \cite{Adam}, Adagrad \cite{Adagrad}, Adadelta \cite{Adadelta}, RMSprop \cite{RMSprop}), penalty terms, and data aggregation techniques (see bagging \cite{bagging} and boosting \cite{boosting}). The lack of the domain knowledge leads to the research on model-driven networks.

%Our OptNet layers are much more computationally expensive than a linear or convolutional layer and a natural question is to ask what the performance difference is. 
%Mit mit tudunk erről mondani?

Despite the popularity of deep learning, %and the advances of model-driven NNs,
traditional ML algorithms continue to dominate in many 1D signal-processing tasks \cite{serkanarXiv}, especially in biomedical signal classification, for example, of electroencephalograms (EEGs), electromyograms (EMGs), and ECGs. The main reason for this lies in the nature of clinical applications, where both accuracy and explainability are important. These cannot be guaranteed by the previously mentioned NN approaches, since they do not extract medically interpretable features. VPNet, however, breaks this impasse by harnessing the theory of variable projection (VP) to provide a framework for solving nonlinear least-squares problems, whose parameters can be separated into linear and nonlinear ones. In many fields of signal processing, there is a large number of linear parameters, which are driven by a smaller number of nonlinear variables (see Eq.~\eqref{eq:vpfunc}). For example, signal compression, representation, and feature-extraction algorithms are often based on linear coefficients of some transformation, such as Fourier and wavelet transforms, which can be parameterized via properties of the window function, mother wavelet, etc.

%Old version: For example, signal compression, representation, and feature-extraction algorithms are often based on linear coefficients of some transformation, such as Fourier transform, wavelet transform, or eigenvalue decomposition. The corresponding nonlinear parameters have a physical meaning; for example, they can be described by frequencies, properties of the window function or free parameters of the wavelets \cite{genVP}. 

The VPNet was designed to merge the expert knowledge used by traditional model-based approaches with the learning abilities of NNs. The proposed architecture is inspired by the so-called model-driven NN concept, which is a emerging trend in signal processing. In Section~\ref{sec:relwork}, we review the existing literature on incorporating model-based information into machine learning. The theoretical background, the general formulation of VPNet, and the corresponding forward and backpropagation algorithm are discussed in Section~\ref{sec:VPNet}. Section~\ref{sec:experiments} describes multiple experiments we performed to evaluate and compare the performance of VPNet to that of other NNs. Finally, Section~\ref{sec:conc} presents conclusions and the expected broader impact of our research.

\section{Related works}
\label{sec:relwork}

Approximation theory gives a general framework to approach the fundamental task in machine learning that is to learn a good representation of the data~\cite{replearning}. Classical methods in approximation theory build up complicated functions by using linear combinations of elementary functions, whereas neural networks use compositions of simple functions. The structure of these compositions constrains the feasible region where we search for the solution of the corresponding ML task. The model-driven NN concept implements these constraints such that the design of the NN architecture resembles the solution to well understood mathematical problems, such as ordinary or partial differential equations~\cite{neuralODE, deepPDE} (ODE, PDE), signal~\cite{deepunfolding,deepunfolding3, wiener-hammersteinNN} and image~\cite{reactdiff, bilevel, learnPDE} processing, optimization~\cite{optNet}, and control~\cite{dynsysNN, wiener-neural, wiener-neural2}.

ODE- and PDE-constrained learning strategies belong to a family of model-driven ML techniques that relates the rigorous mathematical background of differential equations to deep learning problems. On one hand, numerical solvers provide various ways to derive and to interpret the output of NN architectures, such as residual neural networks~\cite{neuralODE}, Hamiltonian networks~\cite{deepPDE}, based on the discretization scheme of the corresponding ODE and PDE. On the other hand, deep learning can incorporate domain knowledge automatically which would otherwise require a significant human effort~\CH{added citation}\cite{yang_multi-object_2019, learnPDE, reactdiff}, e.g. good insights into the problem, and mathematical formulation of a priori information.  Although this approach does not necessarily reduce the number of trainable weights, it helps to design reversible architectures that allow for memory-efficient implementations~\cite{reversibleDNN}.

%Other 
%Model-driven NN constructions, such as deep unfolding \cite{deepunfolding} and Wiener-, and Hammerstein-type NNs \cite{wiener-hammersteinNN} are attempts to overcome the aforementioned drawbacks of DNNs. 

Another branch of model-driven NNs, such as deep unfolding~\cite{deepunfolding} or Wiener-\cite{wiener-neural2}, and Hammerstein-type~\cite{wiener-hammersteinNN} NNs, originates from signal processing problems.
The former approach unfolds the iterations of classical model-based algorithms into layer-wise NN structures whose parameters are optimized based on the training data. This way the resulting NN retains the powerful learning ability of DNNs, inherits expert knowledge, and reduces the size of the training data \cite{DNNtelecom}. Wiener- \cite{wiener-neural2} and Hammerstein-type \cite{wiener-hammersteinNN} NNs are alternatives that combine the advantages of model-based methods and deep learning techniques. These networks comprise cascades of static nonlinear elements and dynamic linear blocks that represent NNs and linear time-invariant (LTI) systems, respectively. Recently, these methods have shown great potential in many fields, for instance, in system identification \cite{wiener-hammersteinNN}, control engineering \cite{wiener-neural2}, sparse approximation theory \cite{sparseLinInv, neuralISTA}, and telecommunication \cite{unfoldcomm, deepunfoldingRF}. 

The motivation behind integrating optimization problems into DNN architectures is similar to the ODE/PDE-driven networks, namely, designing optimization problems to real-world processes is a labor-intensive work which also needs expert knowledge. To date, several new NN architectures have been proposed in order to learn these optimization problems automatically from data. Solving ill-posed inverse problems is a typical example for such neural networks. In this case, each layer is constrained by a penalized linear least-squares problem where the parameters of the regularization term, such as threshold values, linear kernels, weights of the shrinkage functions, constitute the trainable weights~ \cite{bilevel, shrinkageFields, variationalNN}.      OptNet\cite{optNet} gives the most general framework in this family, where the layers encode convex quadratic programming (QP) problems. 
The Hessian matrix of the QP's objective function along with its equality and inequality constraints are learnable parameters. The representation power of an OptNet layer is higher than that of the two-layer ReLU networks, which can reduce the overall depth of DNN architectures (see Theorems 2 and 3 in Ref.~\refcite{optNet}). Besides its advantages, the forward/backward passes of an OptNet layer are much more computationally expensive than a linear or convolutional layer. This is due to the fact that constrained QP problems have no closed form solution in general, thus the forward pass requires the use of iterative numerical solvers in each layer for each update. \CH{start add some out-of-scope references}
We acknowledge that there are many other model-based~\cite{pereira_fema_2020} and model-free approaches~\cite{lara-benitez_asynchronous_2020, peng_deep_2021, kazi_dynensamble_2020, ahmadlou_enhanced_2010, sanchez-reolid_deep_2020}. Especially, for time series data there are methods based on spiking neural networks~\cite{req_spiking, simp_spiking} including their variations~\cite{lara-benitez_experimental_2021, song_spiking_2021} which are beyond the scope of this paper.
\CH{end}

To the best of our knowledge, this is the first time that the VP operators have been exploited in the context of learning end-to-end systems. However, we note that the proposed VPNet can be considered a special case of OptNet. Indeed, a VP layer forwards the solution of an unconstrained separable nonlinear least-squares (SNLLS) problem to the next layer (cf. Eq.~(1) in Ref.~\refcite{optNet}). The corresponding nonlinear parameters are the trainable weights of the VP layer, and the linear ones are the extracted features, which are forwarded to the next layer. In contrast to a general OptNet layer, both the solution and the gradients of a VP layer can be calculated analytically that is provided by the theoretical framework of variable projection~\cite{golub_pereyra1973}. This speeds up the training and the inference, which can be further improved by the use of orthogonal and discrete orthogonal function systems (see e.g. Section~\ref{sec:vpforwardprop}). 

\section{Variable Projection Networks}
\label{sec:VPNet}

\subsection{Variable projections}

Variable Projection (VP) \cite{golub_pereyra1973} provides a framework for addressing nonlinear modeling problems of the form
\begin{equation}
x \approx \hat{x} = \sum\limits_{k=0}^{n-1} c_k \Phi_k({\theta}) = {\Phi}({\theta}) {c},
\label{eq:vp}
\end{equation}
where $x \in \mathbb{R}^m$ and $\Phi_k \in \mathbb{R}^m$ denote the input data to be approximated and a parametric function system, respectively. The symbol $\Phi(\theta)$ refers to both the function system itself and a matrix of size $\mathbb{R}^{m \times n}$. The linear parameters ${c} \in \mathbb{R}^n$ and the nonlinear parameters ${\theta} \in \mathbb{R}^p$ of the function system ${\Phi}$ are separated. The least-squares fit of this problem means minimization of the nonlinear functional
\begin{equation*} 
r({c},{\theta}) := \left\|x - {\Phi}({\theta}) {c}\right\|_2^2. \label{eq:vpproblem}
\end{equation*}
Without nonlinear parameters (i.e.,\ if ${\theta}$ is fixed), the model is linear in the coefficients ${c}$. The minimization of $r$ with respect to $c$ leads to the well-known linear least-squares approximation. Note that it is in fact the best approximation problem in Hilbert spaces. The optimal solution can be expressed by means of Fourier coefficients and orthogonal projection operators $\mathcal{P}_{{\Phi}(\theta)}$:
\begin{equation}
{c} = {\Phi}^+(\theta) x, \qquad \hat{x} = \mathcal{P}_{{\Phi}(\theta)} x = {\Phi}(\theta){\Phi}^+(\theta) x,
\label{eq:vpop}
\end{equation}
where ${\Phi}^+(\theta)$ denotes the Moore--Penrose pseudoinverse of matrix ${\Phi}(\theta)$. The concept is closely related to mathematical transformation methods, such as Fourier and wavelet transforms, that can be interpreted as orthogonal projections by a given function system with a predefined $\theta$. From a practical point of view, the coefficients $c$ can be interpreted as features extracted by VP, and $\hat{x}$ is a result of low-pass filtering and dimension reduction. The minimization of $r$ in the general case can be decomposed into the minimization by the nonlinear parameters ${\theta}$, while the linear parameters ${c}$ are computed by the orthogonal projection. Thus, according to the work of Golub and Pereyra~\cite{golub_pereyra1973}, minimizing $r$ is equivalent to minimizing   the following VP functional:
\begin{equation}
r_2({\theta}) := \left\|x - {\Phi}({\theta}){\Phi}^+({\theta}) x\right\|_2^2.
\label{eq:vpfunc}
\end{equation}
In Ref.~\refcite{varpro_matlab}, a robust gradient-based Matlab implementation were provided for the numerical optimization of $r_2$. Mathematically, VP is a formalization for adaptive orthogonal transformations that allows filtering and feature extraction by means of parametric function systems. If a nonlinear optimization problem can be separated into linear and nonlinear parameters, VP may also act as a solver, which opens up other possible applications \cite{golub_pereyra2003, SNLLS2021}.

In the ML context, VP can be used as a feature extraction method and as a modeling technique for the training procedure \cite{vptrain}. Pereyra et al.\ proposed VP as an optimization method for a given class of feedforward NNs. They modelled the whole network with VP and used the VP optimization method from Ref.~\refcite{golub_pereyra1973} as an alternative to stochastic gradient methods. This methodology is, however, limited to NNs with only one hidden layer. Approaching VP from a different and novel direction, based on its feature extraction ability, we introduce VPNet. 

Previous results have shown that several biomedical signal-processing problems can be addressed efficiently with variable projection by means of adaptive rational and Hermite functions as well as B-splines \cite{genVP, ensembleECG}. VP features have been used in particular for ECG and EEG representation, compression, classification, and segmentation \cite{ECGdelin,weightedHermite,ecgtalk,qrsmodel,ECMI2018,tbme_paper,ratECG,ratECGseg,ratECGclass}. The results show that VP provides a very compact, yet morphologically accurate, representation of signals with respect to the target problem. Additionally, the nonlinear parameters themselves carry direct morphological information about the signals, and they are usually human-interpretable.

\subsection{VPNet architecture}
The key idea of this architecture is to create a network that combines the representation abilities of VP and the prediction abilities of NNs in the form of a composite model. The basic VPNet architecture is a feedforward NN, where the first layer(s) applies a VP operator that is forwarded to a fully connected, potentially deep NN (see Fig.~\ref{fig:vpnet}). The construction is similar to that of CNNs in the sense that the first layer(s) of the network can be interpreted as a built-in feature extraction method. Note that more complex VPNet architectures are also possible, for instance, based on the models of U-Net \cite{ronneberger2015u} and AutoEncoder \cite{Goodfellow-et-al-2016}, which will be investigated as part of our future work.

\begin{figurehere}
\begin{center}
    \includegraphics[width=0.48\textwidth]{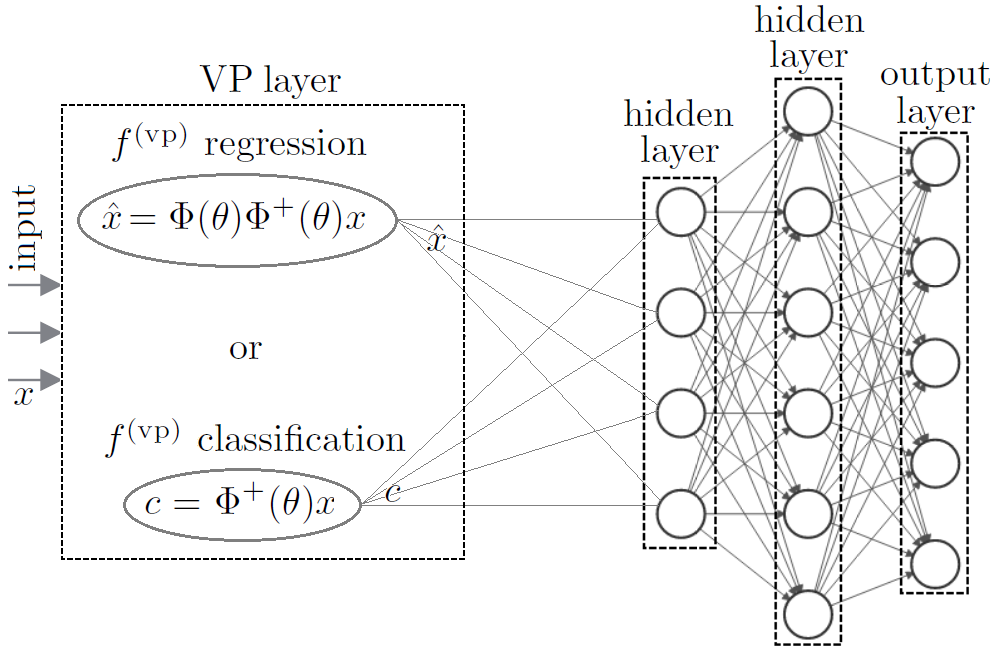}
    \caption{VPNet architecture.}
    \label{fig:vpnet}
\end{center}
\end{figurehere}

Depending on its target application, the VP layer we propose has two possible behaviours. It either performs a filtering of the form
\begin{equation}
f^{\text{(vp)}}(x) := \Phi(\theta)\Phi^+(\theta) x = \hat{x} \qquad (x \in \mathbb{R}^m),
\label{eq:vplayerfilt}
\end{equation}
or a feature extraction of the form
\begin{equation}
f^{\text{(vp)}}(x) := \Phi^+(\theta) x = c \qquad (x \in \mathbb{R}^m),
\label{eq:vplayerfeat}
\end{equation}
where ${\theta} \in \mathbb{R}^p$ denotes the nonlinear system parameters of the given function system ${\Phi}$, as defined above. These VP operators refer to the orthogonal projection and the general Fourier coefficients of the input $x$ by means of the parametric system $\Phi(\theta)$, as in Eq.~\eqref{eq:vpop}. The filter method may be better suited to regression problems, while the feature extraction is suitable for classification problems. The nonlinear system parameter vector ${\theta}$ comprises the learnable parameters of the VP layer.
Note that many inverse problems~\cite{golub_pereyra2003} can be viewed as SNLLS data fitting problems including a small set of adjustable nonlinear parameters ${\theta}$ with direct physical interpretations. For instance, the function system $\Phi_k(t;\tau_k,\lambda_k)=\cos(\lambda_k t + \tau_k)$ can be used in frequency estimation and in EEG, where the network would learn dominant frequencies $\lambda_k$ and phases $\tau_k$ that characterize a certain class of signals, such as seizures in EEG recordings \cite{EEGseizureclass, strf1, strf2}. MRI imaging is another setting \cite{varpro_mri}, where $\Phi_k(t;\lambda_k)=\exp(-\lambda_k t)$ with $\lambda_k\in\IR^+$ yields information about the tissue type. The previously mentioned properties and advantages of the VP operator are implicitly built into VPNet:
%Due to its construction, we expect that the VPNet inherits the previously mentoned properties and advantages of the VP operator: 
%Therefore, VPNet can be configured for a variety of problem domains by choosing an application-specific function systems and a suitable parametrization. 

%According to Ref.~\refcite{golub_pereyra1973}, the gradients of the VP operators can be directly calculated, and thus backpropagation of the VP layer can be done in a natural way, as discussed in Section~\ref{sec:vpbackprop}. 
%Trainable weights: $\theta=(\tau,\lambda)$
%\noindent The properties and advantages of VPNet are:
\begin{itemize}
\item \emph{Role}: A novel model-driven network architecture for 1D signal-processing problems.
\item \emph{Generality}: VPNet can be built from arbitrary parameterized function systems, which allows the direct incorporation of domain knowledge into the network.
\item \emph{Interpretability}: The VP layer can be explained as a built-in feature-extraction method. Further, the layer parameters are the nonlinear VP system parameters, which have an interpretable meaning. They are usually directly connected to morphological properties of the input data (see, e.g.,\ Section~\ref{sec:hermite}).
\item \emph{Simplicity}: Since the VP layer is usually driven by only a few system parameters, VPNet may provide a compact alternative to CNNs and DNNs. In fact, the VP layer can significantly decrease the number of parameters in a DNN.
%the number of parameters are significantly less than in a DNN.
\end{itemize}

\subsection{VP forward propagation}
\label{sec:vpforwardprop}
In order to calculate the forward pass of the VP layer, a linear least-squares (LLS) problem has to be solved for a certain value of $\theta$ in each training iteration (see Eqs.~\eqref{eq:vplayerfilt}-\eqref{eq:vplayerfeat}). Several numerical methods exist to solve such problems, among which QR factorizaton and singular value decomposition (SVD) are the most common techniques. The QR method (requires $\sim 2mn^2 - 2n^3/3$ flops) is fast and reliable for well-conditioned problems, but may fail when ${\Phi}({\theta})\in\IR^{m\times n}$ is nearly rank-deficient. Therefore, in our implementation, we utilize the SVD (requires $\sim 2mn^2 + 11 n^3$ flops) that is the most stable way to solve unconstrained LLS problems~\cite{numlinalg}. Although, it is computationally more demanding than the QR factorization in cases when $m\sim n$, their complexity is approximately the same if $m\gg n$. Note that the latter inequality usually holds in practice, since in VPNet $m$ stands for the length of the input signal, which is much greater than the number of extracted features $n$.

\CH{start complexity}
The low computational complexity is based on the fact that the non linearity is precomputed and stored in the ${\Phi}$-matrix. 
As a consequence, during evaluation, the VP layer just performs a matrix multiplication.
Further, since the number of features computed by the VP layer is typically very low, the following layers can have lower complexity as well.
The weight matrix of a fully connected layer, following the VP layer, is element of $\IR^{n\times l}$ instead of $\IR^{m\times l}$ without the VP layer, with $n$ is the number of coefficients, $m$ is the length of the input signal and $l$ is the number of neuron in the fully connected layer.
Since, $n$ is usually by far smaller than $m$, the weight matrix is significantly smaller for a fixed number or neuron $l$.
\CH{end}

For shallow neural networks, when only a few hidden layers are connected to the VP layer, solving the corresponding LLS problem in each training iteration is obviously the bottleneck of VPNet that influences both the computational complexity and the numerical accuracy. In the following, we provide a realization of the VP layer with Hermite functions, and we demonstrate how the choice of the function system and its parametrization influence the conditionality of ${\Phi}({\theta})$. 

\subsubsection{Adaptive Hermite system}
\label{sec:vpforwardprop_Hermite}
In order to alleviate the computational burden of the VP layer, a straightforward option is to parametrize orthogonal function systems. As a case study, let us consider Hermite polynomials \cite{szego}, which are defined by the three-term recurrence relation: 
\begin{equation*}
	H_{k+1}(t) = 2tH_k(t)-2kH_{k-1}(t) \quad (k\in\IN^+, t\in \IR)\textrm{,}
\end{equation*}
where $H_0(t)=1$ and $H_1(t)=2t$. These classical orthogonal polynomials can be parametrized via dilation and translation:
\begin{equation}
	%\Phi_k(\tau,\lambda):=
	\Phi_k(t;\tau,\lambda)=\sqrt{\lambda} \Phi_k(\lambda (t-\tau)),
	\label{eq:adapthsys}
\end{equation}
where
\begin{equation}
\Phi_k(t) = H_k(t)e^{-t^2/2}/\sqrt{\pi^{1/2}2^k k!} \qquad (k\in\IN^+)\textrm{.}
\label{eq:hermitefun}
\end{equation}
The functions $\Phi_k(t;\tau,\lambda)$ are the translated and dilated variations of the well-known Hermite functions, thus we refer to them as "adaptive Hermite functions". 

The forward propagation of the corresponding Hermite-VP layer can be defined by the matrix $\Phi(\theta)$ in Eq.~\eqref{eq:vpfunc}. For a given parameter value $\theta=(\tau,\lambda)$, the $k$-th column of $\Phi(\theta)$ is equal to the values of the $k$-th adaptive Hermite function evaluated at some predefined points $t_0, t_1,\ldots,t_{m-1}\subseteq[a;b]$, where $[a;b]$ stands for the sampling interval. In the case of proper discretization \cite{gautschi}, the columns of $\Phi(\theta)$ are pairwise orthogonal and unit vectors for all $\theta$; therefore, $\Phi^+(\theta)=\Phi^T(\theta)$, which speeds up the computation of both the forward and the backward passes.   

There are two strategies for choosing the discretization points: nonuniform and uniform sampling. The former relies on the Gauss--Hermite quadrature rules, which associates the points $t_0, t_1,\ldots,t_{m-1}\subseteq[a;b]$ with the roots of Hermite polynomials~\cite{gausshermite}. This approach is the most accurate way to define discrete orthogonal systems, but it requires both the precomputation of the roots and the resampling of the input signals at these nonequidistant points. Therefore, we consider the computationally simpler uniform discretization instead. This sampling scheme, although less accurate, satisfies discrete orthogonality, and thus the identity $\Phi^+(\theta)=\Phi^T(\theta)$ holds, provided that the number of sampling points $m$ is large enough, and $\theta\in\Gamma$, where    
\begin{align*}
	\Gamma= \left\{(\tau,\lambda)\in\IR\times\IR_+\;:\; \tau + \frac{3}{\lambda} \leq b, \quad \tau - \frac{3}{\lambda} \geq a\right\}\textrm{.}
\label{eq:const_affin}
\end{align*}
If $\theta\notin\Gamma$, it can happen that the adaptive Hermite functions $\Phi_k(t,\tau,\lambda)$ are not discrete orthogonal anymore. In the worst-case scenario, they can be linearly dependent, which results in a rank deficient matrix $\Phi(\theta)$. In Fig.~\ref{fig:cond_fun}, we demonstrate this phenomenon by evaluating the condition number of $\Phi(\theta)\in\IR^{m\times n}$ for $m=1000,\ n=3$, and for a range of parameters $\theta=(\tau,\lambda)\in [500,1100] \times [0.05,0.012]$. It can be seen that the condition number diverges from the ideal case (green dashed line) as we change $\tau$ and $\lambda$ irrespective of $\Gamma$. This can be avoided if we choose the parameters from the feasible region $\Gamma$. The rationale behind this behaviour is given in Appendix~A.

\vspace{4mm} %spare some space for the top label
\begin{figurehere}
\centering
	\begin{tikzpicture}
    \begin{axis}[%
   	name=plot1,
   	compat=newest, 
%		unit vector ratio*=0.6 1 1,
		scale=0.8,
		xlabel=$\tau$ translation, 
		ylabel style={align=center}, 
		ylabel style={text width=3.4cm},
		ylabel={condition number}, 
%		scaled x ticks=base 10:-1,
%		scaled y ticks=base 10:6,
		legend pos=north west, 
		legend cell align=left,
		legend columns=1, 
		xmin=500, xmax=1120,
	  ymin=0, ymax=40,
        legend style={at={(0,1)}, anchor=north west,
                    % the /tikz/ prefix is necessary here...
                    % otherwise, it might end-up with `/pgfplots/column 2`
                    % which is not what we want. compare pgfmanual.pdf
            /tikz/column 2/.style={
                column sep=5pt,
            },
        font=\small},
		grid=major]
			\addplot[line width=1pt, color=blue] table[x index =0, y index =2] {Figures/condition_number1.dat};
			\addlegendentry{$\mathrm{cond}(\bs{\Phi}(\tau))$  for a fixed $\lambda$}
			
			\addplot[line width=1pt, color=red] table[x index =0, y index =3] {Figures/condition_number1.dat};
			\addlegendentry{$\mathrm{cond}(\bs{\Phi}(\lambda))$ for a fixed $\tau$}

		  \addplot[line width=1.0pt, color=green, style=dashed] table[x index =0, y index =1] {Figures/condition_number2.dat};
			\addlegendentry{$\mathrm{cond}(\bs{\Phi}(\tau,\lambda))$ for $(\tau,\lambda)\in\Gamma$}

    \end{axis}
    \begin{axis}[%
   	name=plot1,
   	compat=newest, 
%		unit vector ratio*=0.6 1 1,
		scale=0.8,
		axis y line=none,
		axis x line*=top,
		xlabel style={align=center}, 
		xlabel style={text width=3.4cm},
		xlabel={$\lambda$ dilation}, 
		xtick={600, 800, 1000},
		xticklabels={4.37, 3.11, 1.84},
		scaled x ticks=base 10:2,
		legend pos=north west, 
		legend cell align=left,
		legend columns=1, 
		xmin=500, xmax=1120,
	  ymin=0, ymax=40,
		]
		\end{axis}	
\end{tikzpicture}
\caption{Relationship between the parameters $\tau,\lambda$ and the condition number of the matrix $\bs{\Phi}(\tau,\lambda)$.}%
\label{fig:cond_fun}
\end{figurehere}

\subsection{VP backpropagation}
\label{sec:vpbackprop}

Let us discuss the training of a general feedforward NN in a supervised manner. Let
\[ (x_i,y_i) \qquad (i=1,2,\dots,N) \]
be the annotated input-target pairs of the training data, where the input vector $x_i \in \mathbb{R}^m$ and the target vector $y_i \in \mathbb{R}^s$ (in the case of regression) or the target label $y_i \in \mathbb{N}$ or probabilities $y_i \in [0,1]^c$  (in the case of classification). A general feedforward NN can be expressed as the composition of layer functions of the form
\[ NN_{\boldsymbol\theta}(x)=\left(f_{\theta^{(L)}}^{(L)}\circ \ldots \circ f_{\theta^{(\ell)}}^{(\ell)} \circ \ldots \circ f_{\theta^{(2)}}^{(2)} \circ f_{\theta^{(1)}}^{(1)}\right)(x), \]
where $x \in \mathbb{R}^m$ stands for the input samples, $f_{\theta^{(\ell)}}^{(\ell)}$ and $\theta^{(\ell)}$ denote the function and the parameters of layer $\ell$, respectively. The symbol $\boldsymbol\theta$ refers to the set of parameters $\theta^{(\ell)}$. The layer functions $f^{(\ell)}$ may refer to linear mappings, convolutional filters, nonlinear activations, pooling, VP operators, etc. Let
\[ \hat{y}_i := NN_{\boldsymbol\theta}(x_i) \qquad (i=1,2,\dots,N) \]
denote the predicted values for each input. The training of the network can be addressed as a minimization problem, involving a proper loss (i.e., cost) function $J$ that evaluates the error between predicted and target values. Common loss functions are the Mean Squared Error (MSE), that is, the least-squares cost function (regression problems, $y_i \in \mathbb{R}^s$), and the Binary Cross Entropy (BCE) loss (binary classification, $y_k \in \{0,1\}$):
\begin{align*}
    J_{MSE}(\boldsymbol\theta) &:= \dfrac{1}{N}\sum\limits_{i=1}^N \|y_i - \hat{y}_i \|_2^2, \\ J_{BCE}(\boldsymbol\theta) &:= -\dfrac{1}{N}\sum\limits_{i=1}^N \left(y_i \log \hat{y}_i + (1-y_i) \log(1 - \hat{y}_i) \right). 
\end{align*}    
In our experiments, we used the Cross Entropy loss $J_{CE}$, which is the multi-class extension of BCE (classification, $y_k \in \mathbb{N}$); see also Section~\ref{sec:experiments}.

The state-of-the-art method for training feedforward networks is backpropagation~\CH{added citation}\cite{rumelhart_learning_1987}, where $J$ is minimized by means of a stochastic gradient-descent optimization (see e.g. Adam~\cite{Adam}, Adagrad~\cite{Adagrad}, RMSprop~\cite{RMSprop}). 
\CH{start add some backprop infos / citations}
There are multiple implementation of the propagation algorithm for different programming languages, target hardware platforms and machine learning frameworks~\cite{hung_parallel_1993, hung_oo_backprop_1994, paszke2017automatic, tensorflow2015-whitepaper}. 
\CH{end}
The gradient descent update formula for each layer parameter is
\[ \theta^{(\ell)} := \theta^{(\ell)} - \eta \dfrac{\partial J}{\partial \theta^{(\ell)}}, \]
where $\eta > 0$ is called the learning rate. Briefly, backpropagation provides a recursive way of computing the gradients above based on the chain rule:
\[ \dfrac{\partial J}{\partial f^{(\ell-1)}} = \dfrac{\partial J}{\partial f^{(\ell)}} \cdot \dfrac{\partial f^{(\ell)}}{\partial f^{(\ell-1)}}, \qquad \dfrac{\partial J}{\partial {\theta}^{(\ell)}} = \dfrac{\partial J}{\partial f^{(\ell)}} \cdot \dfrac{\partial f^{(\ell)}}{\partial {\theta}^{(\ell)}}. \]
This way, only the partial derivatives of the layer function $f^{(\ell)}$ with respect to its input $(\partial f^{(\ell)}/\partial f^{(\ell-1)})$ and to its parameters $(\partial f^{(\ell)}/\partial {\theta}^{(\ell)})$ must be calculated. These derivatives are usually well known for the common layer types and can also be directly calculated for the VP layers. Based on Ref.~\refcite{golub_pereyra1973}, the partial derivatives of the VP operators with respect to their input and nonlinear parameters can be expressed as follows. In the case of a filtering-type VP layer (see Eq.~\eqref{eq:vplayerfilt}):
\begin{align*}
    f^{\text{(vp)}}(x) &= \Phi(\theta)\Phi^+(\theta) x, \qquad \dfrac{\partial f^{\text{(vp)}}}{\partial x} = \left[\Phi(\theta)\Phi^+(\theta)\right]^T, \\ \dfrac{\partial f^{\text{(vp)}}}{\partial \theta_j} &= \dfrac{\partial \left[\Phi(\theta)\Phi^+(\theta)\right]}{\partial \theta_j} x,
\end{align*}    
where
\[ \partial\left[\Phi(\theta)\Phi^+(\theta)\right] = (I - \Phi\Phi^+) \partial \Phi \Phi^+ + \left[(I - \Phi\Phi^+) \partial \Phi \Phi^+\right]^T.\]
In the case of a feature-extraction-type VP layer (see Eq.~\eqref{eq:vplayerfeat}):
\begin{align*}
f^{\text{(vp)}}(x) &= \Phi^+(\theta) x, \qquad \dfrac{\partial f^{\text{(vp)}}}{\partial x} = \left[\Phi^+(\theta)\right]^T, \\ \dfrac{\partial f^{\text{(vp)}}}{\partial \theta_j} &= \dfrac{\partial \Phi^+}{\partial \theta_j} x, 
\end{align*}
where
\begin{align*}\partial \Phi^+ = &-\Phi^+ \partial \Phi \Phi^+ + \Phi^+ \left[\Phi^+\right]^T \partial \Phi^T (I - \Phi\Phi^+) \\ &+ (I-\Phi^+\Phi) \partial \Phi^T \left[\Phi^+\right]^T \Phi^+.
\end{align*}

The naive implementation of the backpropagation, particularly in the case of DNNs, can lead to numerical issues, such as divergence and overfitting.
%Instead of a fixed learning rate, a better practice is the utilization of decaying or the aforementioned adaptive gradient methods. Besides the variation of the learning rate, a common technique is the regularization by means of penalty terms. For instance, a simple regularization of a linear layer is the $\ell_2$ penalty on the weight parameters that may reduce the overfitting effect \cite{Goodfellow-et-al-2016}. 
In order to avoid this, a regularization term in the form of an $\ell_2$ penalty on the weight parameters is added to the loss \cite{Goodfellow-et-al-2016}. 
Here, we introduce a percent root-mean-square difference (PRD) regularization that can be applied to a single feature-extraction VP layer in the case of a classification problem. The modified loss function we propose is
\begin{align*} 
&J_{VP}(\boldsymbol \theta) := J_{CE}(\boldsymbol \theta) + \dfrac{\alpha}{N}\sum\limits_{i=1}^N \dfrac{r_2(x_i; \theta^{\text{(vp)}})}{\|x_i\|_2^2} = \\ &J_{CE}(\boldsymbol \theta) + \dfrac{\alpha}{N}\sum\limits_{i=1}^N \dfrac{\left\|x_i - \Phi\left(\theta^{\text{(vp)}}\right)\Phi^+\left(\theta^{\text{(vp)}}\right) x_i\right\|_2^2}{\|x_i\|_2^2}, 
\end{align*}
where $\alpha \ge 0$ controls the penalty effect. The motivation behind this regularization is twofold: First, it is based on the previous results that incorporate VP as feature extraction, which show that the precise VP approximation may lead to 'good' features and therefore to high classification accuracy. Second, we expect that the optimal VPNet classifier extracts the main characteristics of the input signals, which means that we presume 'good' approximation. This penalty term seemingly breaks the formulation of the backpropagation, but the original method can easily be extended by a bypass step that is applied to the VP layer only. The gradient with respect to the VP parameters is modified as follows:
\[ \dfrac{\partial J_{VP}}{\partial \theta^{\text{(vp)}}} = \dfrac{\partial J_{CE}}{\partial \theta^{\text{(vp)}}} + \dfrac{\alpha}{N}\sum\limits_{i=1}^N \dfrac{1}{\|x_i\|_2^2} \cdot \dfrac{\partial r_2}{\partial \theta^{\text{(vp)}}}, \]
where
\[ \partial r_2 = -2 x_i^T (I-\Phi\Phi^+) \partial \Phi \Phi^+ x_i. \]
We just developed the formulas for attaining the necessary gradient information for training VPNet via backpropagation. This allows us to train VPNets in the same way as convolutional and fully connected NNs.

\section{Experiments}
\label{sec:experiments}

Using supervised classification problems inspired by particular biomedical signal-processing applications, we evaluated VPNet and compared it to fully connected and 1D convolutional networks. We present the details of the experiments, specifically about the network architectures, the VP system of choice, and the synthetic and real datasets.

\subsection{Network architecture}
\label{sec:architecture}

Here we provide details about the networks we compared, the learning methods, and the network parameters. The networks were feedforward, consisting of the following layers:
\begin{itemize}
\item \emph{VPNet}: a VP layer, a fully connected (FC) layer with ReLU activation, an FC layer with SoftMax activation;
\item \emph{Fully connected NN}: one or two FC layers with ReLU, an FC layer with SoftMax;
\item \emph{CNN}: a 1D convolutional and pooling layer, an FC layer with ReLU, an FC layer with SoftMax.
\end{itemize}

For signal-classification tasks, the inputs were $\mathbb{R}^m$ samples and the outputs were interpreted as a probability distribution over predicted output classes. The FC layers performed linear mappings with nonlinear activation (ReLU or SoftMax). The VP layer was of the feature-extraction type (see Eq.~\eqref{eq:vplayerfeat}), and the CNN implemented 1D convolution and mean or maximum pooling as in Ref.~\refcite{serkan2016}.

Based on cross entropy loss with VP regularization (see Section~\ref{sec:vpbackprop}), offline backpropagation with Adam optimizer \cite{Adam} was applied for learning. The hyperparameters and the parameter selection strategies were as follows:
\begin{itemize}
\item \emph{Learning parameters}: learning rate, VP penalty (VPNet only), batch size, and the number of epochs. The last two were fixed (512 and 10-100, respectively). The optimal learning rate and penalty can be found by a grid search.
\item \emph{Network parameters}: number of layers, number of neurons, VP dimension $n$ (VPNet only), convolutional and pooling kernel sizes (CNN only). Here we either used fixed dimensions so that the three architectures are comparable or evaluated possible configurations by a grid search.
\item \emph{Layer parameters}: linear weights and biases, nonlinear VP parameters (VPNet only), kernel weights and biases (CNN only). These parameters were optimized by backpropagation. Initialization was random for the linear and kernel parameters. However, the VP parameters have interpretable meaning, which may lead to special initialization. We investigated two options: a grid search on the intervals of possible values and initialization by means of pretraining the VP layer to reconstruct input data (i.e., minimizing $r_2$ in Eq.~\eqref{eq:vpfunc}). The latter approach is especially useful in the case of complex waveforms which possibly need more VP parameters to learn.  
%when the search space of the VP layer is large, i.e., it includes several nonlinear parameters.  (e.g., in the case of complex waveforms).  
\end{itemize}

\begin{figure*}[!t]
\begin{center}
    \includegraphics[width=0.97\textwidth]{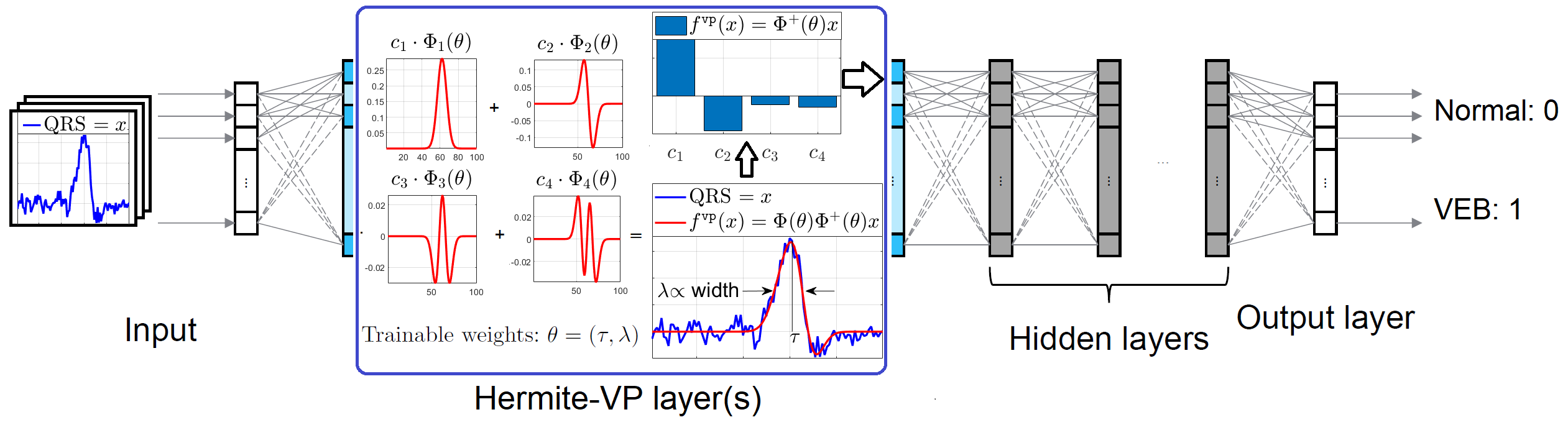}
    \caption{VPNet architecture for QRS classification: the VP layer takes the whole signal as input, decomposes the QRS complexes into linear combinations of adaptive Hermite functions, and then forwards the coefficients of the Hermite components to the next fully connected layer.}
    \label{fig:hermitevpnet}
\end{center}
\end{figure*}

\subsection{VP system of choice}
\label{sec:hermite}
Although Hermite functions have shown great potential in many fields, such as molecular biology \cite{molbio}, computer tomography \cite{ctapp}, radar \cite{hermiteradar}, and physical optics \cite{optics}, their main application area is 1D biomedical signal processing. The shape features of Hermite functions are well suited to producing models of compactly supported waveforms such as spikes \cite{hexp8, hexp4, hexp5, hexp1, hexp7}, which is why we used them in ECG heartbeat classification. 

The nonlinear parameters $\tau$ and $\lambda$ in Eq.~\eqref{eq:adapthsys} represent the time shift and the width of the modeled waveforms, respectively. Thus, the network learns the positions and the shapes of those waves/spikes which separate one class from another. For instance, in electrocardiography, a heartbeat signal comprises three individual waveforms (i.e.,\ the QRS, T, and P waves), which represent different phases of the cardiac cycle, and their properties are directly used by medical experts for diagnosis. These features are learned by the VP layer: The amplitude and shape information is extracted by the linear coefficients $c_k$, while position and width of the waves are represented by $\tau$ and $\lambda$ (see Fig.~\ref{fig:hermitevpnet}). This approach is essentially different from CNN-based methods, where no direct connections exist between learned and medical descriptors.

%A detailed review of these applications and their VP reformulation can be found in \cite{golub_pereyra2003}.
%(see e.g.\ the review in \cite{golub_pereyra2003}).

%According to \cite{golub_pereyra2003}, several applications can be reformulated as a VP problem. 
%Többi alkalmazást be lehet tenni a broader impacthoz: radar, EMG, gait classification, neural spike classification, stb.

\subsection{Synthetic data}

Our goal was, on the one hand, to synthesize a dataset where we know the actual structure of the data depending on the generator parameters. On the other hand, the dataset had to have practical relevance (i.e., be related to actual signal-processing problems). The generator system of choice was the adaptive Hermite system, which seemed to fulfill these expectations due to its applications in signal processing (see Section~\ref{sec:hermite}). The principles we followed to generate the dataset are discussed below.

Let us consider a general signal model by means of a linear combination of adaptive Hermite functions of the form
\[
x_i = \Phi(\tau_i, \lambda_i) \cdot c^{(i)} = \sum\limits_{k=0}^{n-1} c^{(i)}_{k} \Phi_k(\tau_i, \lambda_i),
\]
where $(\tau_i, \lambda_i)$ and $c^{(i)}$ $(i = 1,2,\dots,M)$ refer to the sample-specific nonlinear parameters and coefficients, respectively. Based on the completeness of the Hermite system in $L^2(\mathbb{R})$, this formula provides a general approximation for arbitrary signals. However, the signal-processing applications of VP and the Hermite system show that proper selection of the nonlinear parameters may lead to accurate low-order approximations. Further investigation into this topic revealed that the nonlinear parameters correspond to coarse changes in the signal morphologies, while the coefficients reflect fine details~\cite{tbme_paper2}. For instance, we refer to Ref.~\refcite{ensembleECG}, where the nonlinear parameters where utilized as global, patient-specific and the coefficients as heartbeat-specific descriptors. Motivated by these aspects, we sought to construct a dataset where the nonlinear parameters are close to each other and the coefficients form noticeably separable classes. 

\begin{figure*}[!t]
\vspace{-2em}
%\subfloat[][Coefficients]{\includegraphics[width=0.33\textwidth]{./Figures/synhermite_coeffs.pdf}}
\subfloat[][Coefficients]{\includegraphics[width=0.33\textwidth, trim=90 270 90 200, clip]{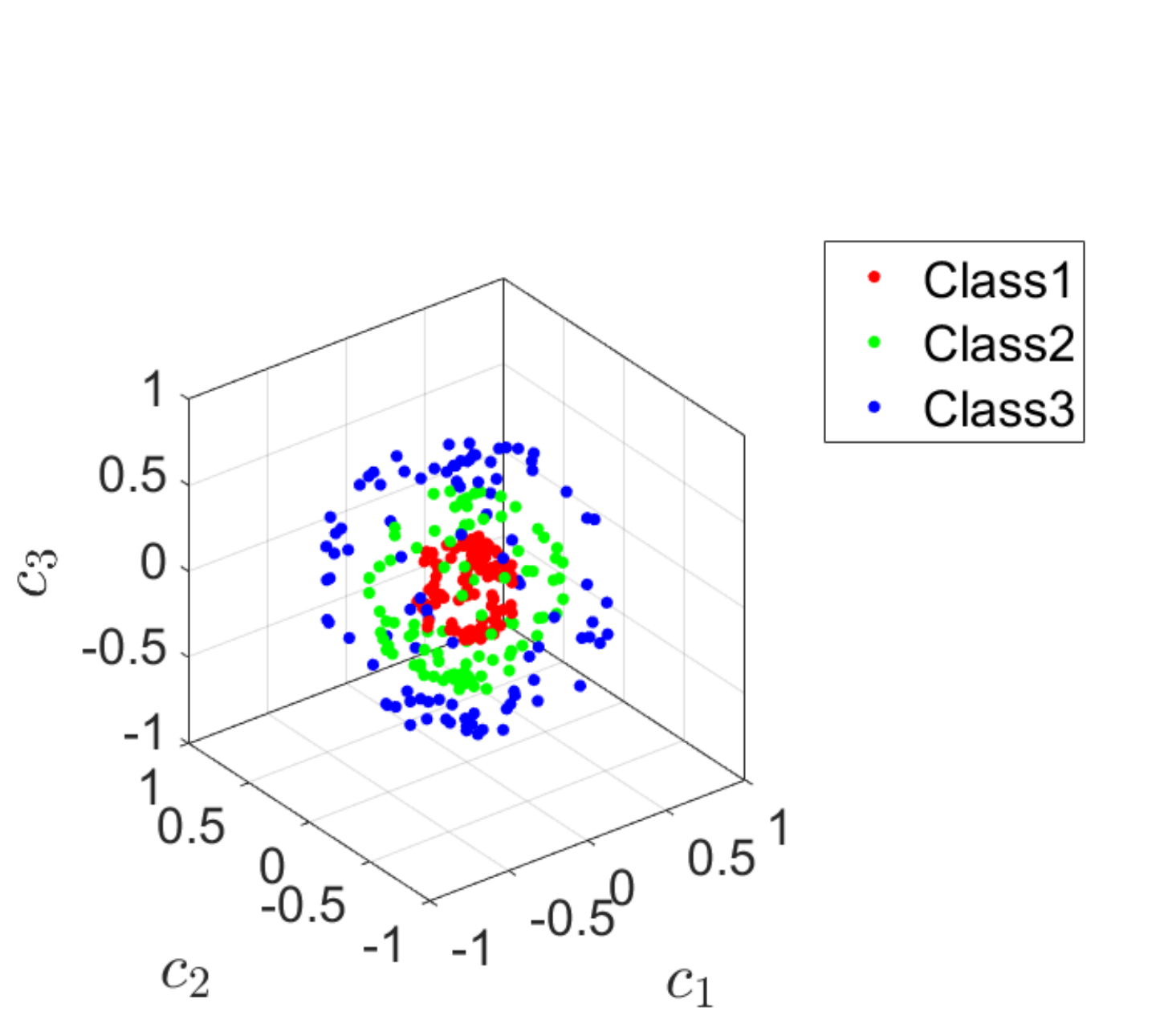}}
\hfil
%\subfloat[][System parameters]{\includegraphics[width=0.33\textwidth]{./Figures/synhermite_params.pdf}}
\subfloat[][System parameters]{\includegraphics[width=0.33\textwidth, trim=90 270 90 200, clip]{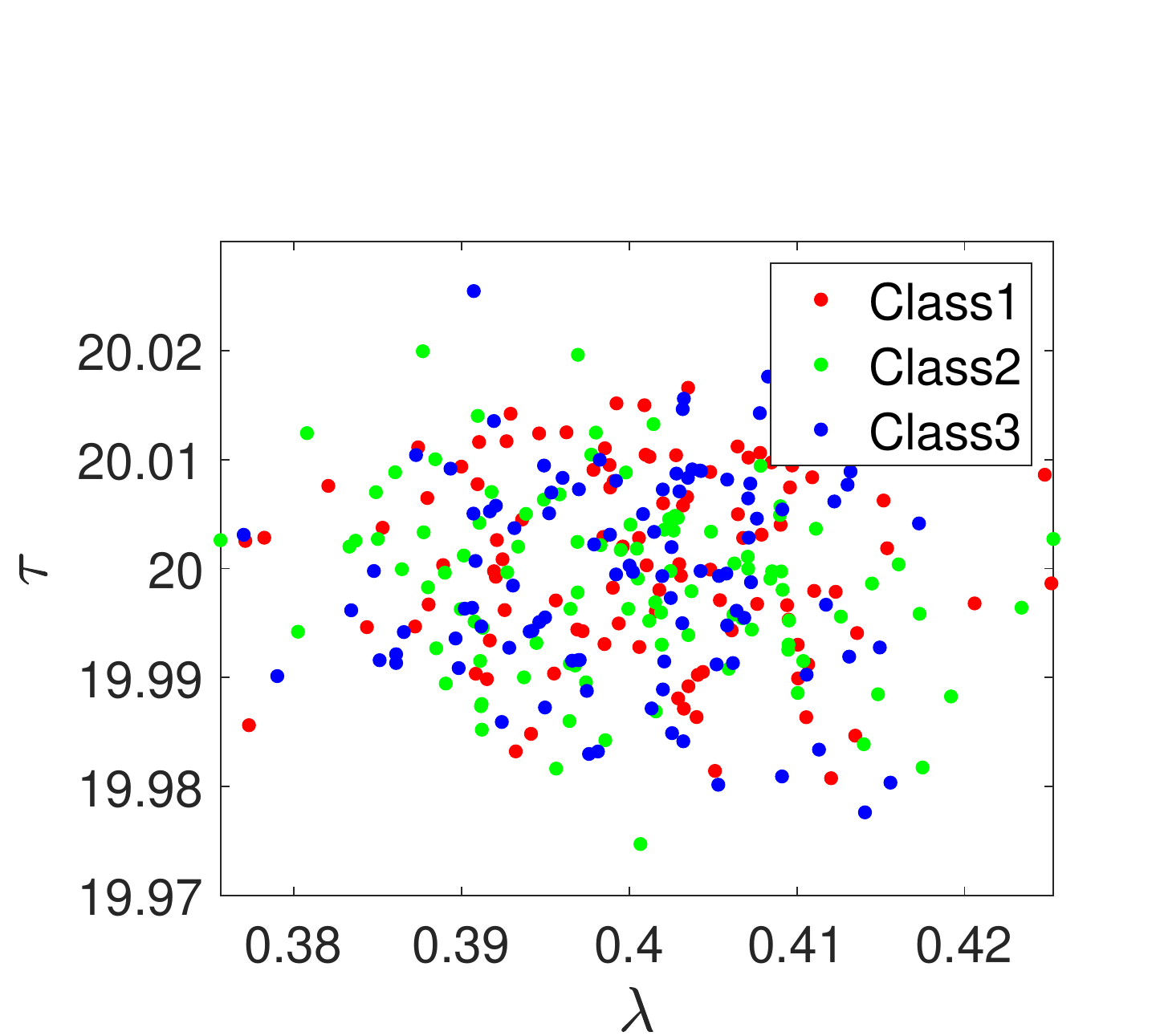}}
\hfil
\subfloat[][Samples]{\includegraphics[width=0.33\textwidth, trim=90 270 90 200, clip]{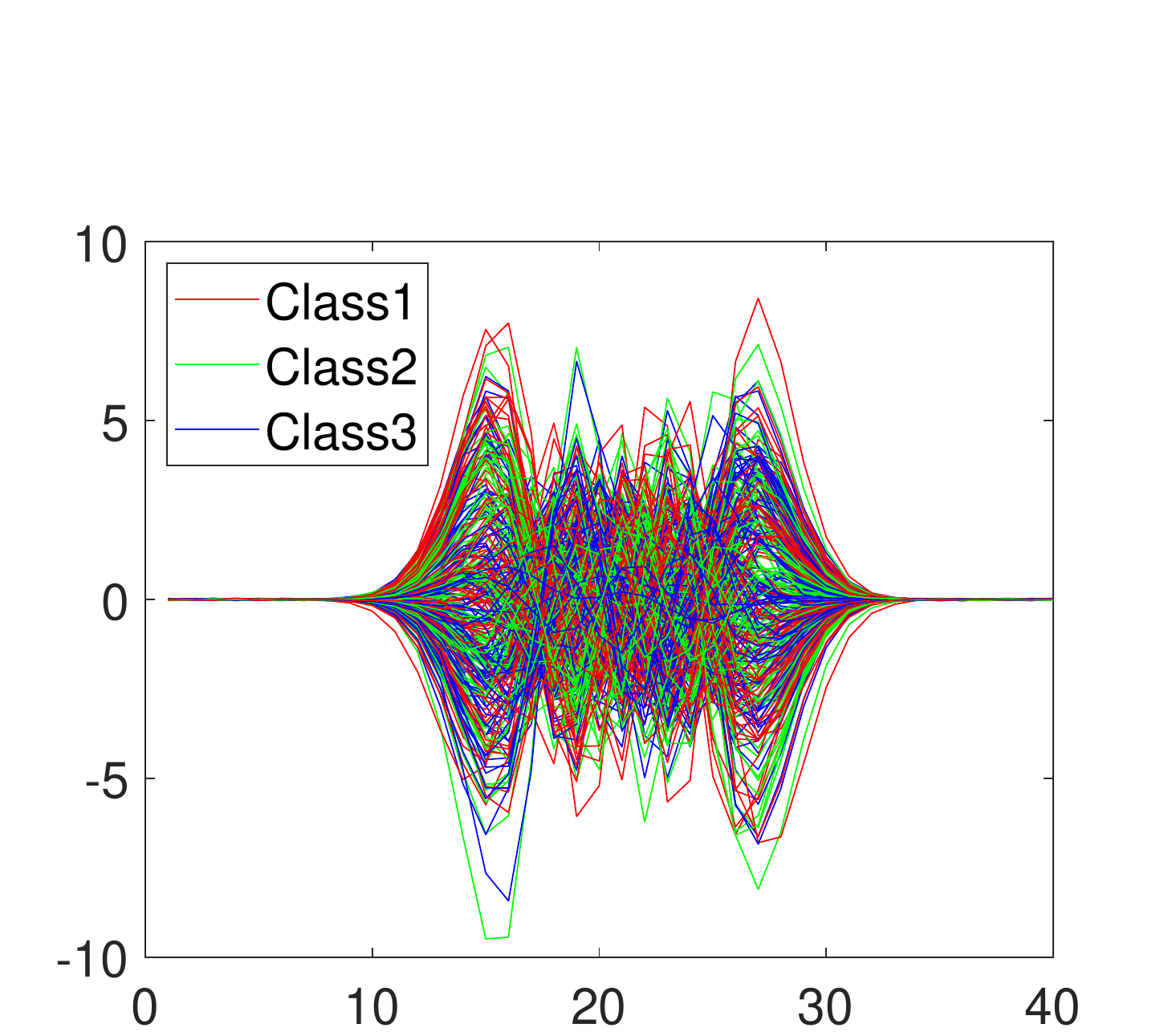}}
\caption{Synthetic dataset: first, the coefficients~(a) and the parameters of the Hermite functions~(b) are generated; which are then used to compute the input samples~(c).}
\label{fig:synhermite}
\end{figure*}

More precisely, we considered five coefficients (i.e.,\ $c^{(i)} \in \mathbb{R}^5$) so that the points $(c^{(i)}_{1},c^{(i)} _{2},c^{(i)}_{3}) \in \mathbb{R}^3$ formed three separable spherical shells that correspond to the target class labels (see Fig.~\ref{fig:synhermite}~(a)). The motivation behind spherical shells was twofold. They are simple enough for human interpretation, but sufficiently complex to require complex networks. The last two coefficients, $c^{(i)}_{4}$ and $c^{(i)}_{5}$, served as random factors and for amplitude normalization. Their effect is to mislead the classifier, but at the same time to decrease the chance of overfitting. The nonlinear parameters $\tau_i$ and $\lambda_i$ are similar for each sample up to a random factor, and the sample-specific parameter values are generated randomly with given mean and variance (see Fig.~\ref{fig:synhermite}~(b)). This random factor simulates the nonlinear noise in the measurement. Fig.~\ref{fig:synhermite}~(c) presents the samples. We conclude that the simulation met our expectations: the resulting
samples were hard to separate, but the underlying
structure was easy to interpret. Note that this is a standard process to generate synthetic data which was utilized by other authors as well~\cite{datagen}.

\begin{figure*}[!t]
\vspace{-2em}
\subfloat[][Best training curves]{\includegraphics[width=0.33\textwidth]{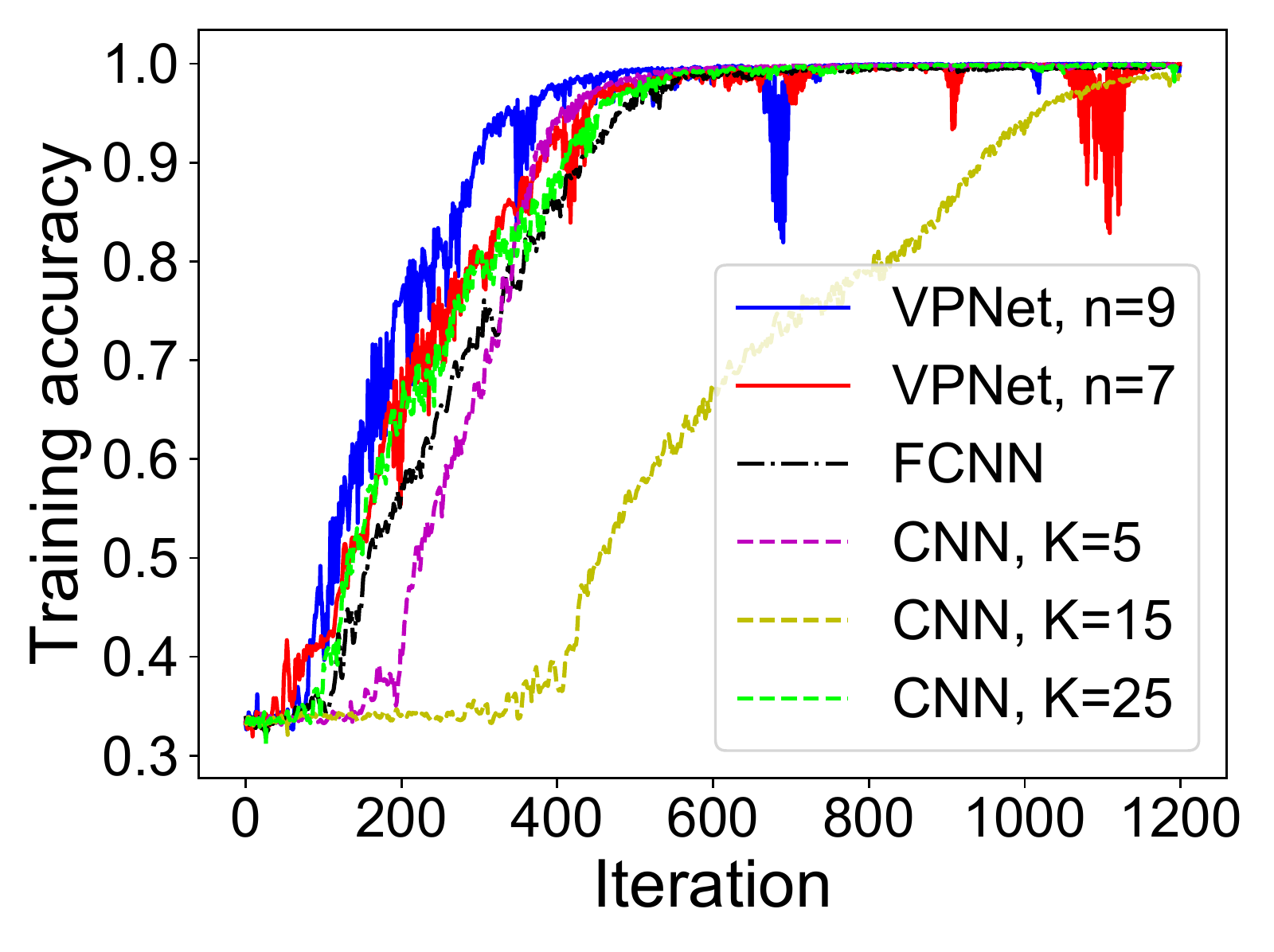}}
\hfil
\subfloat[][Best test accuracies]{\includegraphics[width=0.33\textwidth]{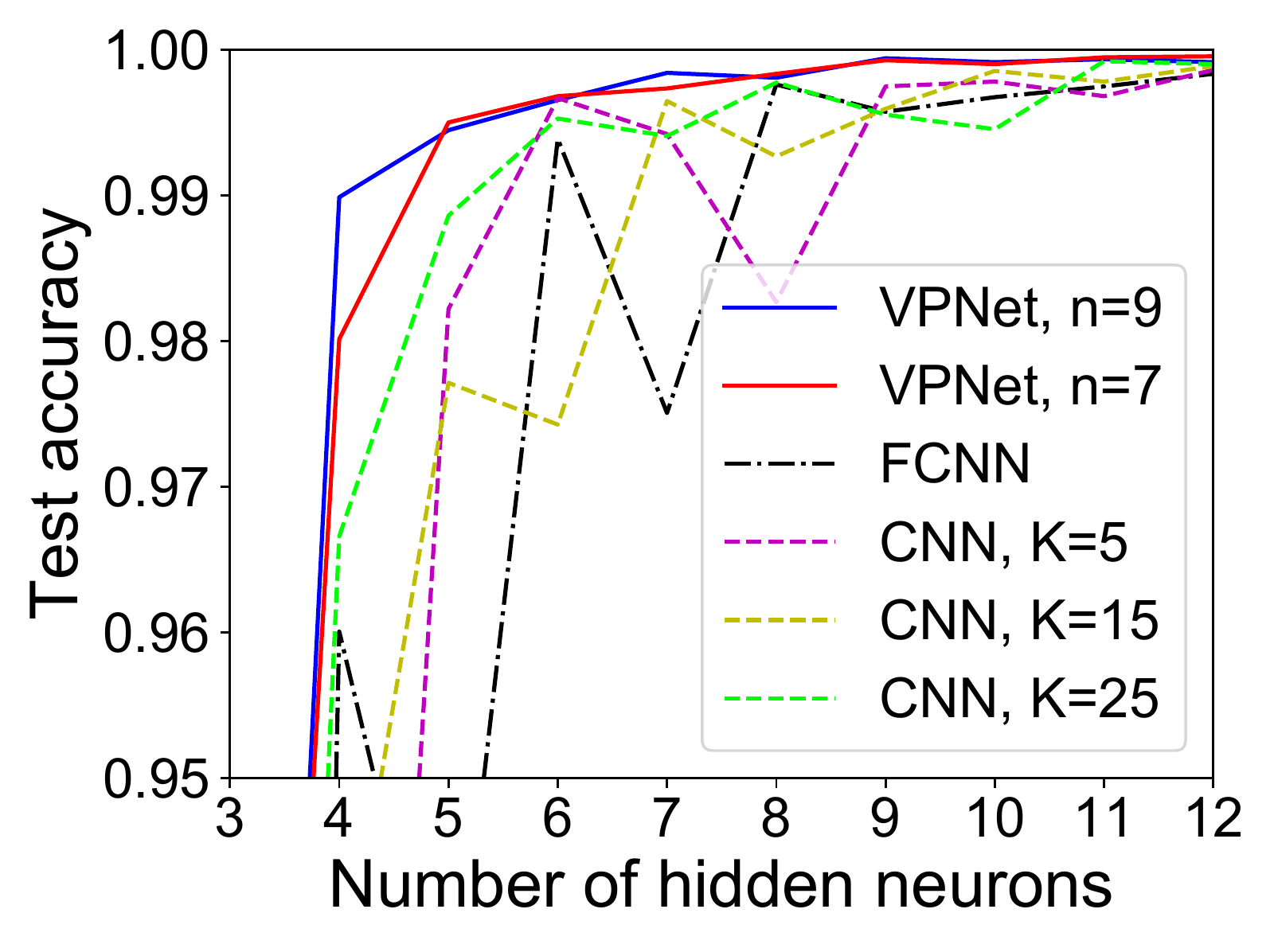}}
\hfil
\subfloat[][Best test accuracies]{\includegraphics[width=0.33\textwidth]{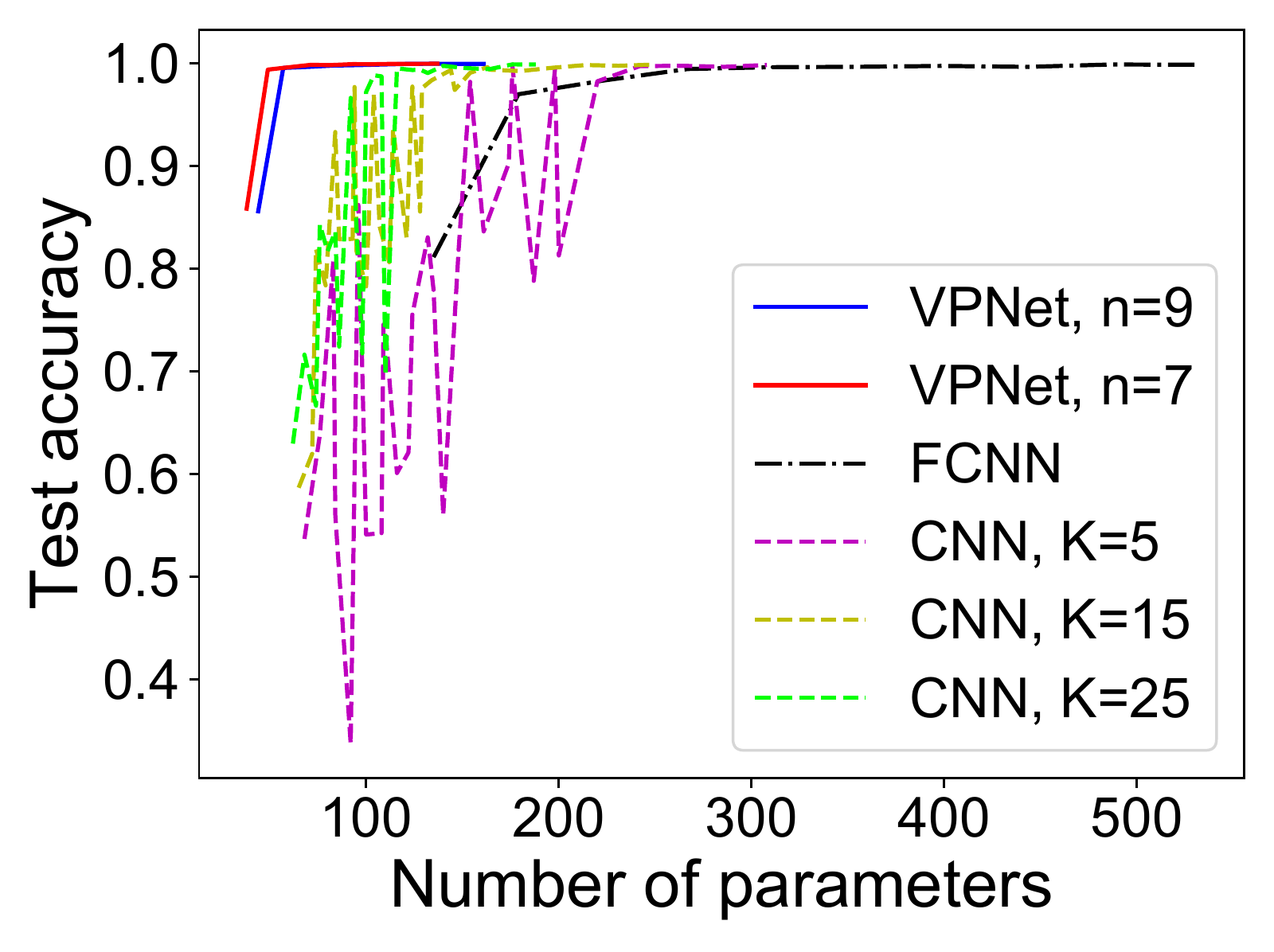}}
\caption{Evaluation on synthetic data}
\label{fig:results}
\end{figure*}

In the actual implementation, 5000 samples per class were generated for both the training and test sets. We evaluated a total of more than 8000 possible hyperparameter configurations of the three network architectures. A range of numbers of neurons in the hidden layer, various numbers of VP dimensions, and various CNN kernel and pooling sizes, learning rates and VP initializations were considered. The VP penalty was initially fixed to $0.1$. The simulations showed that the VP regularization can not only increase the learning speed, but also ensure convergence of an otherwise divergent configuration. In this regard, $0.1$ was found to be a good choice. The aggregated results are presented in Fig.~\ref{fig:results} (a) and (b). There, the configurations are grouped into six categories: VPNets of dimension $n=7$ and $n=9$ in Eq.~\eqref{eq:vp}, fully connected NNs (FCNN), and CNNs with kernel sizes of 5, 15, and 25. Fig.~\ref{fig:results} (a) shows the training accuracy curves corresponding to the best hyperparameter combination in each category. In Fig.~\ref{fig:results}~(b) and (c), the best test accuracies are plotted against the number of neurons in the hidden layer and the total number of learnable network parameters, respectively, for each category. We note that the $y$-axis of Fig.~\ref{fig:results}~(b) is restricted to the interval between $95\%$ and $100\%$ for better visual interpretability. In the following, we compare the performance of VPNet with respect to different network complexities.

The results demonstrate the efficiency and potential capabilities of VPNet. Fig.~\ref{fig:results}~(a) indicates its fast learning ability. In fact, VPNet may converge faster than the other network architectures. Fig.~\ref{fig:results}~(b) and (c) show that VPNet can potentially outperform FCNNs and CNNs in terms of the best accuracies on the test set. Although all architectures achieved accuracies close to 100\%, VPNet achieved this with low structural complexity, which refers not only to the number of neurons, but also to the total number of network parameters (see Fig.~\ref{fig:results}~(c)). In this regard, VPNet is superior, because with FCNNs and CNNs of the same effective receptive field, the number of parameters (i.e., the linear and kernel weights and biases) grows linearly with sample size and number of neurons. With VPNet, in contrast, the number of nonlinear parameters ($p=2$) is independent of sample size and output dimension. For the sake of clarity, we remark that the kernel size or the number of convolutional layers in a CNN do not necessarily depend on the input size. Although, in order to detect global morphologic behavior of signals (e.g. heartbeats), the CNN is expected to have a large enough effective receptive field, that requires larger kernels or multiple layers stacked together in a linear scale. See also Ref.~\refcite{cnnerf}.

\begin{tablehere}
\tbl{Evaluation on synthetic data: best test accuracies vs. number of parameters\label{tab:results}}
{\begin{tabular}{lccc}
\toprule
\textbf{\#} & \textbf{VPNet} & \textbf{CNN} & \textbf{FCNN} \\
\colrule
30-39 & 85.86\% &  & \\
40-49 & 99.41\% &  & \\
50-59 & 99.57\% &  & \\
60-69 & 99.64\% & 71.65\% & \\
70-79 & 99.87\% & 84.32\% & \\
80-89 & 99.85\% & 93.33\% & \\
90-99 & 99.94\% & 97.71\% & \\
100-119 & 99.97\% & 98.86\% & \\
120-139 & 99.98\% & 99.41\% & 81.14\%\\
140-159 & 99.97\% & 99.77\% & \\
160-179 & 99.96\% & 99.92\% & 97.01\%\\
180-199 &  & 99.90\% & \\
200-239 &  & 99.85\% & 98.34\%\\
240-279 &  & 99.89\% & 99.47\%\\
280-319 &  & 99.86\% & 99.65\%\\
320-359 &  &  & 99.68\%\\
360-399 &  &  & 99.77\%\\
400-479 &  &  & 99.67\%\\
480- &  &  & 99.91\%\\
\botrule
\end{tabular}}
\end{tablehere}

In addition to Fig.~\ref{fig:results}~(c), the best test accuracies depending on the number of learnable parameters are given in Table~\ref{tab:results}. Here, the number of parameters are grouped into bins for easier interpretation. The results show that the VPNet outperforms the CNNs and FCNNs for each bin, and reaches peak performance earlier than the other two. Besides the numerical comparison, statistical hypothesis testing were also performed for each bin, if applicable. The differences between the best performing VPNets and CNNs are statistically significant by both paired-sample $t$-tests and McNemar's tests with significance level 5\%.

%\begin{figure*}[!t]
%\vspace{-2em}
%\subfloat[][Normal beats]{\includegraphics[width=0.5\textwidth]{./Figures/ecg_train_n.pdf}}
%\hfil
%\subfloat[][Ventricular ectopic beats %(VEBs)]{\includegraphics[width=0.5\textwidth]{./Figures/ecg_train_v.pdf}}
%\caption{Example heartbeats of the training set}
%\label{fig:heartbeats}
%\end{figure*}

\subsection{Real ECG data}

We sought to prove the relevance of VPNet not only in simulation, but also using real signal-processing data. We chose a particular ECG signal-processing problem: classification of heartbeat arrhythmia (see Ref.~\refcite{aami}). The state of the art is supervised ML by traditional approaches (see Refs.~\refcite{ECGsurvey,ECGsurvey2,ECGsurvey3} and Section~\ref{sec:relwork}), including VP-based static feature extraction \cite{ensembleECG,ratECG,ratECGclass}. Here we focused on a related subproblem where we could compare the performance of the selected network configurations.
%Here we did not aim to outperform these methods. Instead, we focused on a related subproblem where we could compare the performance of the selected network configurations.

In more detail, we investigated the separation of the two largest arrhythmia classes: normal and ventricular ectopic beats (VEBs). The source of the data is the benchmark MIT-BIH Arrhythmia Database \cite{MIT-BIH}, available from PhysioNet \cite{PhysioNet}. The database is split into sets DS1 and DS2 according to Ref.~\refcite{deChazal}, for training and inference, respectively. The whole database contains more than 100~000 annotated heartbeats, but it is heavily biased towards the normal class, that usually distorts the performance evaluation. Here, we investigated two cases for data acquisition. First, a balanced subset was extracted: all VEBs and the same number of normal beats from each record. This yielded 4260 plus 4260 heartbeat signals for training (set DS1), and 3220 plus 3220 signals for testing (set DS2). This balanced subset is expected to provide undistorted evaluation and fair comparison of the NN architectures. The second, unbalanced subset consists of all normal beats and VEBs of the whole database, yielding around 50~000 heartbeats for both training and testing. This unbalanced subset represents a more realistic scenario, and supports partial comparability to the state-of-the-art. Note that the DS1 and DS2 heartbeats come from different patients, which means that there is no data leakage in either cases. We used the preprocessing and heartbeat extraction methods discussed in Ref.~\refcite{ratECG}, but chose a window size of 100 samples ($\sim$ 0.28 s) around the R peak annotations. This window was expected to cover the whole QRS complex and potentially the PR and ST segments of each heartbeat. Example heartbeats of the two classes are displayed in Fig.~\ref{fig:heartbeats}.

\begin{figurehere}
\begin{center}
\subfloat[][Normal]{\includegraphics[width=0.5\linewidth]{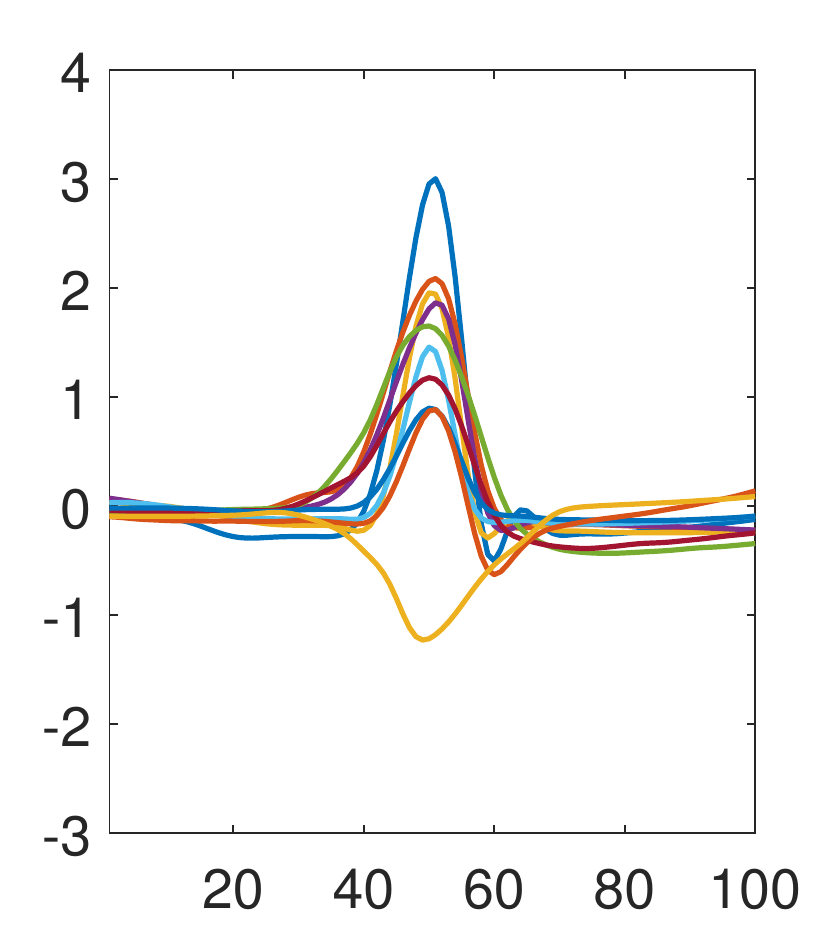}}
\subfloat[][VEB]{\includegraphics[width=0.5\linewidth]{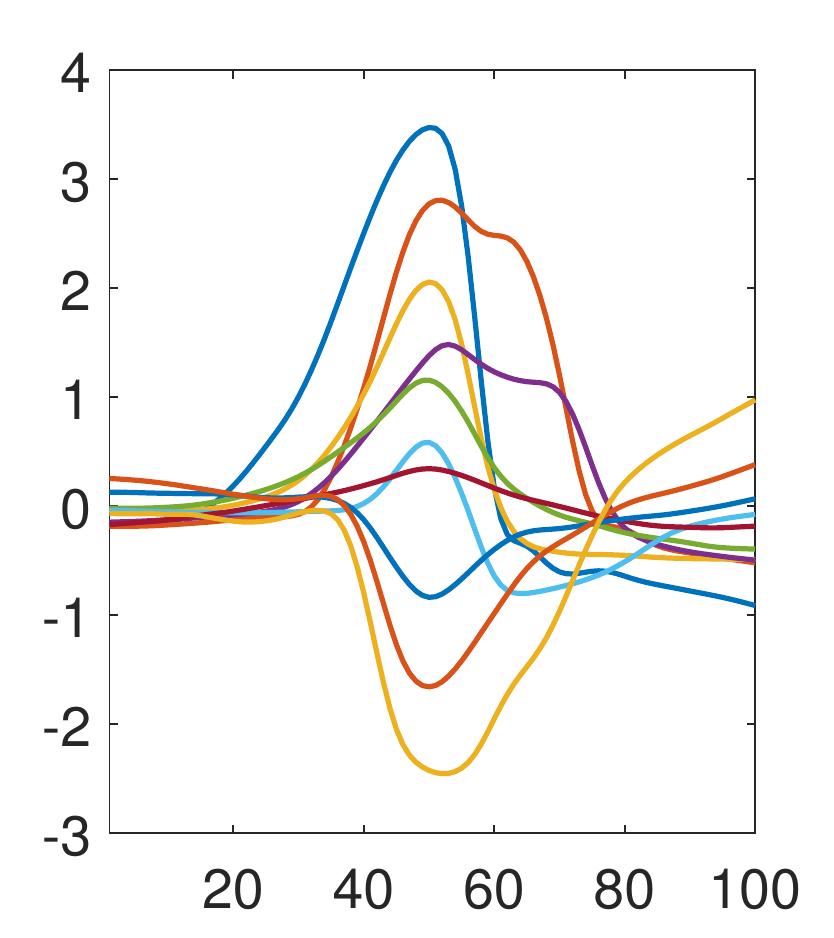}}
\caption{Example heartbeats of the training set}
\label{fig:heartbeats}
\end{center}
\end{figurehere}

\begin{figure*}[!t]
\vspace{-2em}
%\subfloat[][Coefficients]{\includegraphics[width=0.33\textwidth]{./Figures/synhermite_coeffs.pdf}}
\subfloat[][Normal beat]{\includegraphics[width=0.33\textwidth, trim=90 270 90 283, clip]{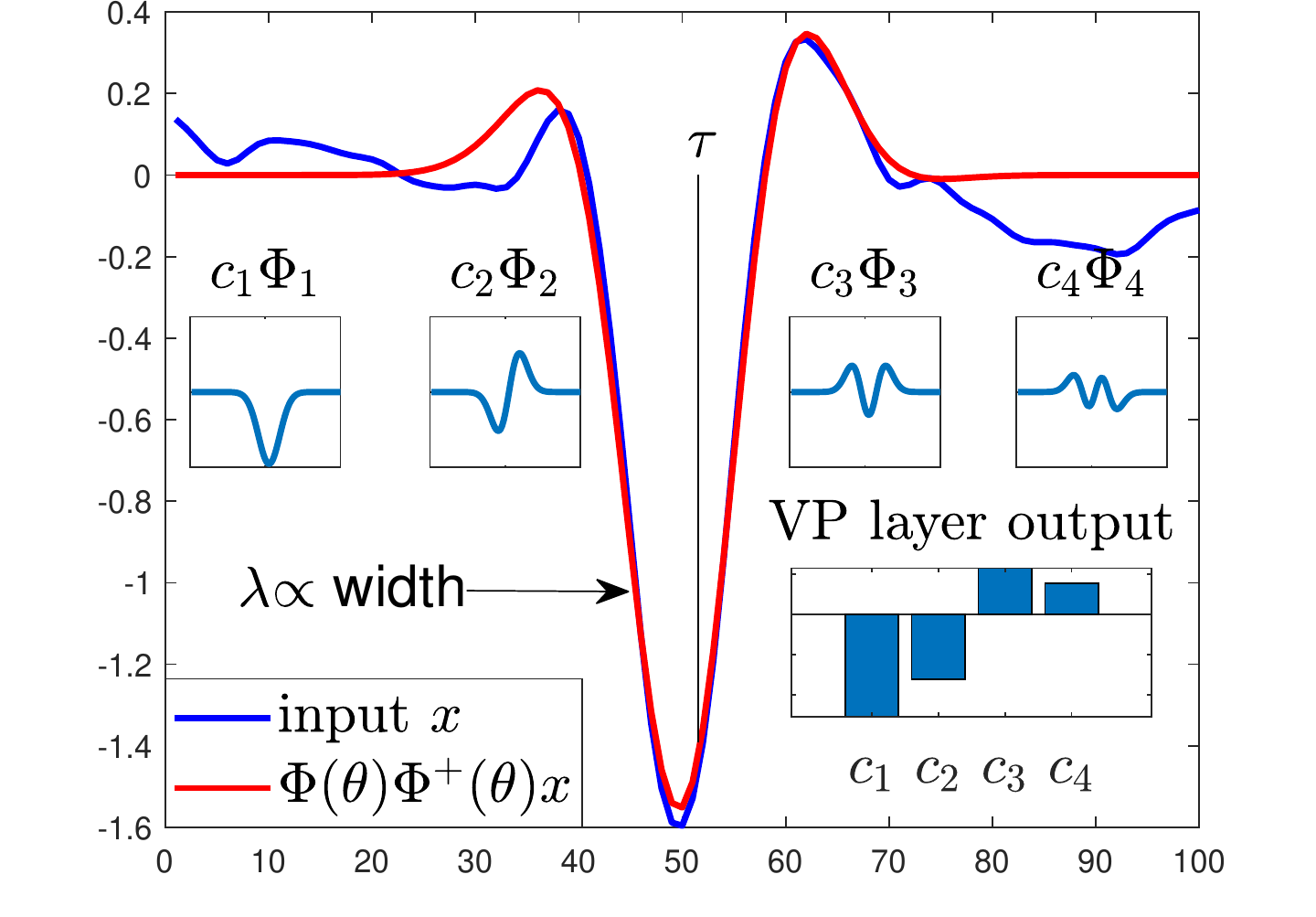}}
\hfil
%\subfloat[][System parameters]{\includegraphics[width=0.33\textwidth]{./Figures/synhermite_params.pdf}}
\subfloat[][Abnormal beat]{\includegraphics[width=0.33\textwidth, trim=90 270 90 283, clip]{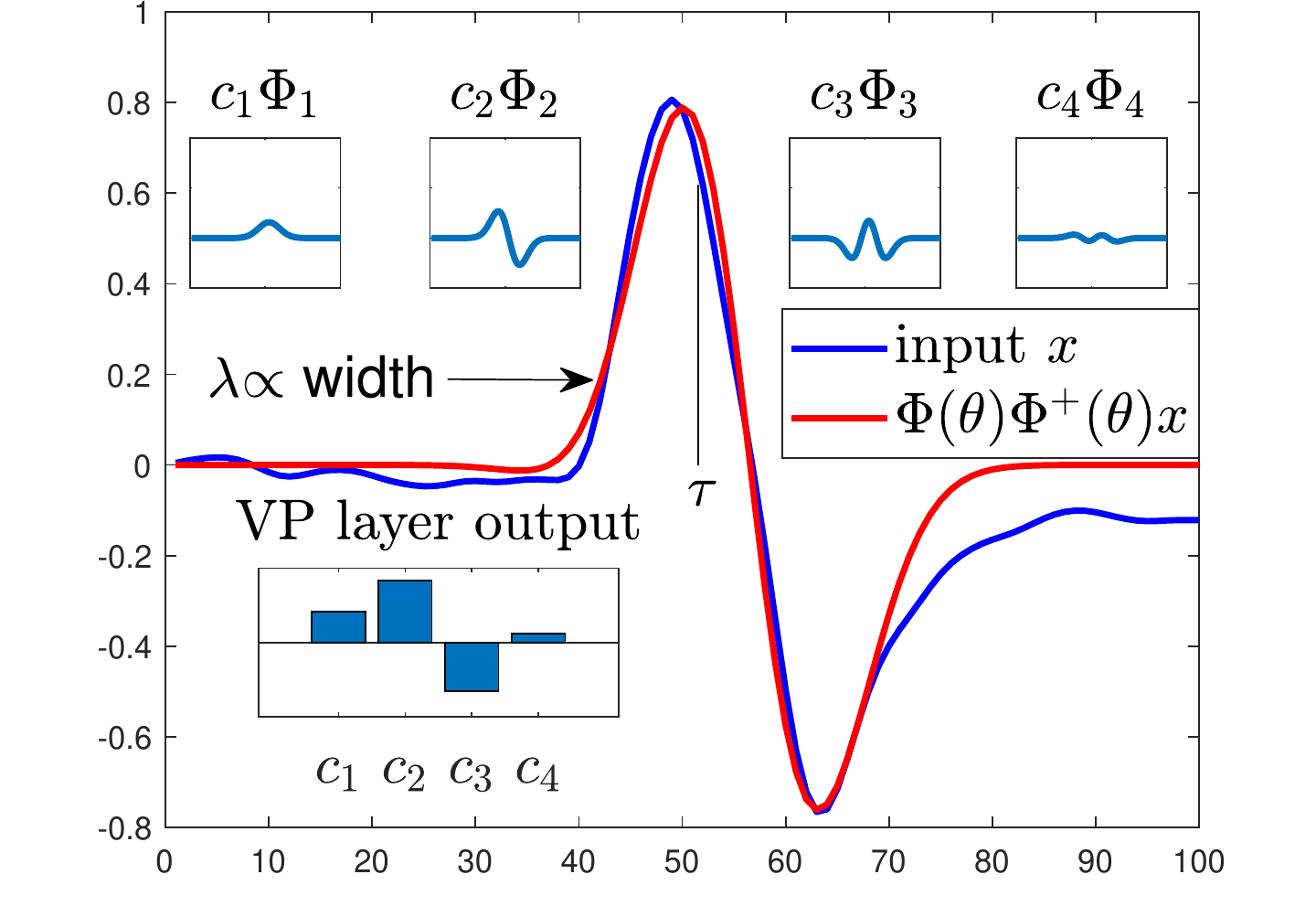}}
\hfil
\subfloat[][Abnormal beat]{\includegraphics[width=0.33\textwidth, trim=90 270 90 283, clip]{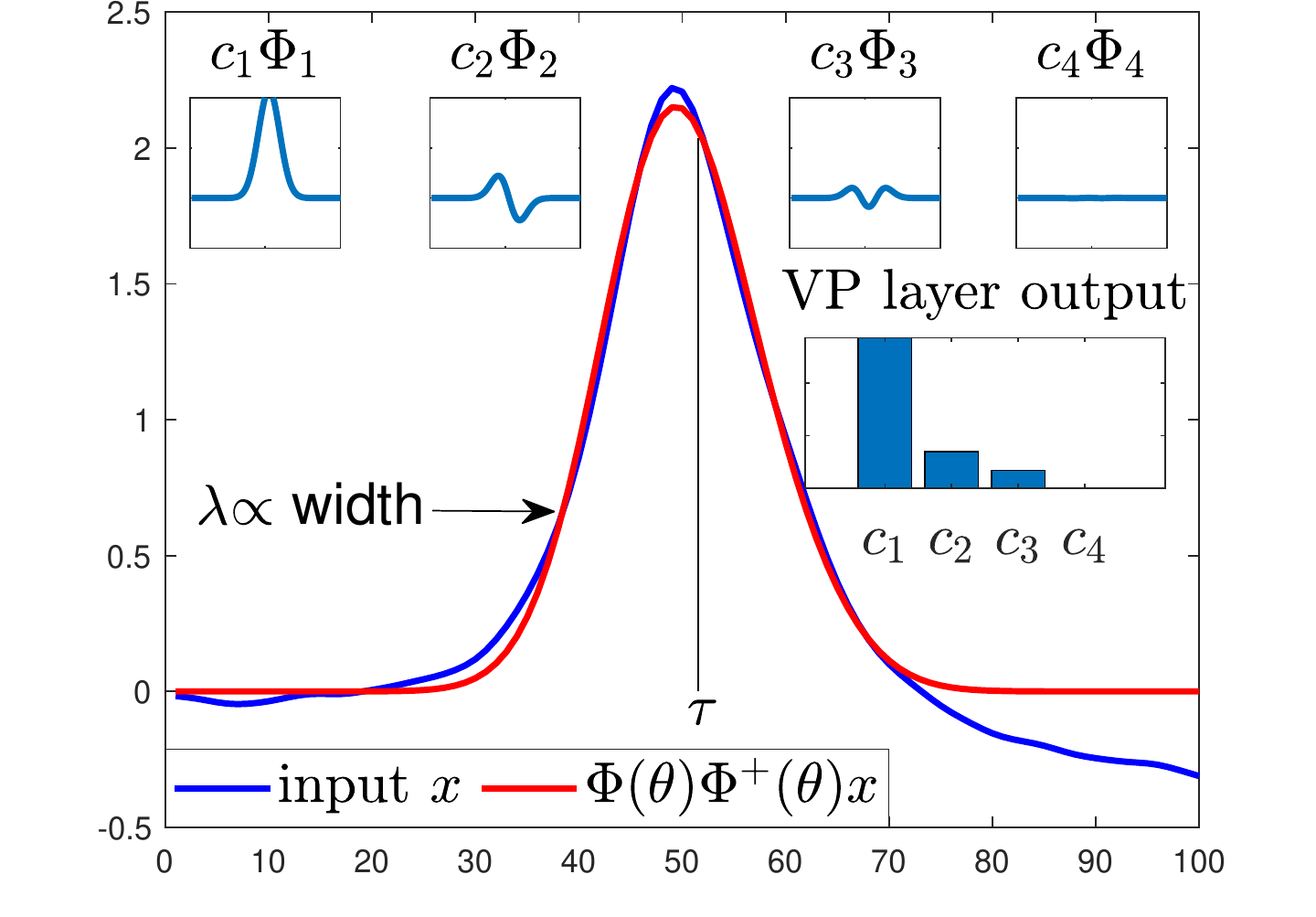}}
\caption{Output of a trained VP layer: for a normal beat~(a) and two abnormal beats~(b)-(c).}
\label{fig:interpret}
\end{figure*}

%In Section~\ref{sec:hermite}, we claimed that the Hermite-VPNet learns the position $\tau$ and the width $\lambda$ of the waves/spikes, and extracts shape information, which is forwarded to the next layer. 
To demonstrate the interpretability of the results, we depicted the response of a trained VP layer to three input QRS complexes in Fig.~\ref{fig:interpret}. It can be seen that the Hermite-VP layer learned in fact the position $\tau$ and the width $\lambda$ of the QRS complexes such that it gives an approximation (red) to the meaningful part of the original (blue) curves. In addition to the QRS complex, the input data window may include irrelevant information, such as baseline wander, noise, part of the P and the T waves. However, these irrelevant information are discarded due to the optimization of $\tau$ and $\lambda$, and thus only the meaningful part of the input signal is approximated at the end of the training. Consequently, the VP layer is likely to be more tolerant to noise as well. In fact, the Hermite-VP representation of ECG recordings can simultaneously cope with
various noise sources such as baseline wander, and power-line interference~\cite{tbme_paper2}. The layer can also retain diagnostically important morphological information via the extracted coefficients. In Fig.~\ref{fig:interpret}, the red curve is equal to the linear combination of the Hermite functions, whose coefficients are the output of the VP layer. The magnitude of these coefficients indicate the presence of each elementary components in the signal. For instance, Fig.~\ref{fig:interpret}~(b) shows an asymmetric QRS complex, which is reflected in a high coefficient $c_2$ that corresponds to an odd Hermite function. In contrast, Fig.~\ref{fig:interpret}~(c) plots a highly symmetric QRS complex, which resembles to a Gaussian function indicated by the high value of $c_1$. Therefore, both the parameters $\tau,\ \lambda$ and the output $c_i$'s of the VP layer are interpretable. Note that the level of interpretability tends to decrease as we connect more and more hidden layers to the network. 
\CH{add a sentence here interpreterbility} 
The reason behind is that the whole network does not seek to reconstruct the parameters with which the data where constructed but it rather searches for the parameters that maximize the distinctness of the classes.
Since the term presented in~\ref{sec:vpbackprop} penalizes the model for not reconstructing the original signal, a larger value for $\alpha$ mitigates the decreased interpretability.
\CH{end}
However, the VP layer provides a fully transparent feature extractor, which directly influences the output of the network due to the least-squares penalty in the modified loss function $J_{VP}$. Therefore, a trained VP layer can be used to improve the generalization properties of DNNs by synthesizing more realistic data samples in the learned feature space~\cite{dataaug, feataug}. 

\begin{table*}[!b]
\tbl{Performance evaluation on real data\label{tab:ecg_results}}
{\begin{tabular}{ccccccc}
\toprule
\multicolumn{2}{c}{\multirow{2}{*}{\textbf{Case/Method}}} & \textbf{Total} & \multicolumn{2}{c}{\textbf{Normal}} & \multicolumn{2}{c}{\textbf{VEB}} \\
&& \textbf{accuracy} & \textbf{$Se$} & \textbf{$+P$} & \textbf{$Se$} & \textbf{$+P$} \\
\colrule
\multirow{3}{*}{Balanced} & VPNet & 96.65\% & 99.38\% & 94.23\% & 93.91\% & 99.34\%\\
& FCNN & 94.38\% & 93.79\% & 94.91\% & 94.97\% & 93.86\% \\
& CNN & 96.34\% & 97.76\% & 95.05\% & 94.91\% & 97.70\%\\
\colrule
\multirow{3}{*}{Unbalanced} & VPNet & 98.45\% & 99.57\% & 98.78\% & 83.07\% & 93.37\%\\
& FCNN & 97.49\% & 98.50\% & 98.81\% & 83.70\% & 80.21\% \\
& CNN & 98.35\% & 99.39\% & 98.85\% & 84.07\% & 90.93\%\\
\colrule
\multicolumn{2}{c}{\textit{State-of-the-art} \cite{ECGsurvey}} & N/A & 80--99\% & 85--99\% & 77-96\% & 63--99\% \\
\botrule
\end{tabular}}
\end{table*}

\begin{figure*}[!t]
\vspace{-2em}
\hfil
\subfloat[][Balanced subset]{\includegraphics[width=0.33\textwidth]{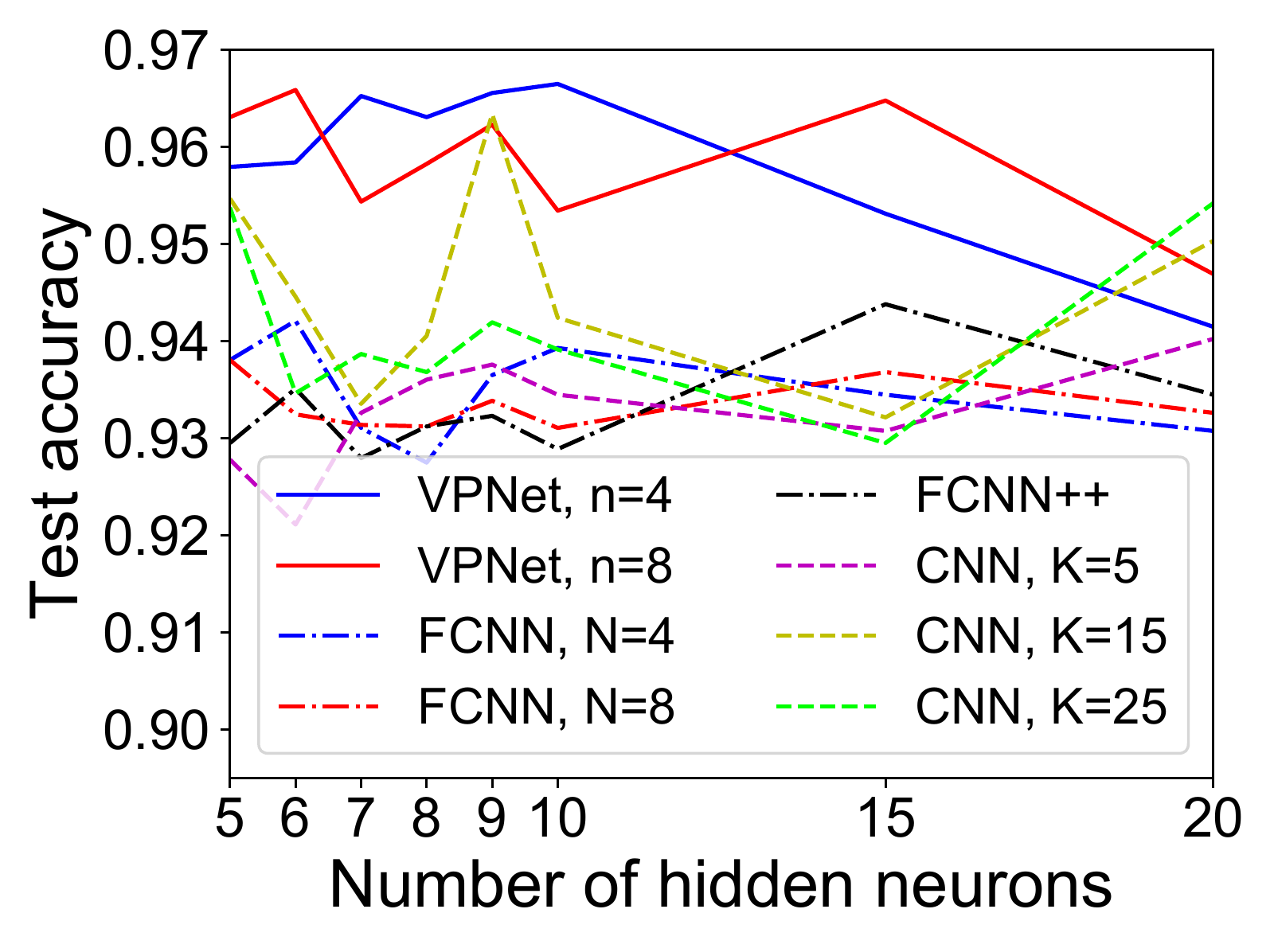}}
\hfil
\subfloat[][Unbalanced subset]{\includegraphics[width=0.33\textwidth]{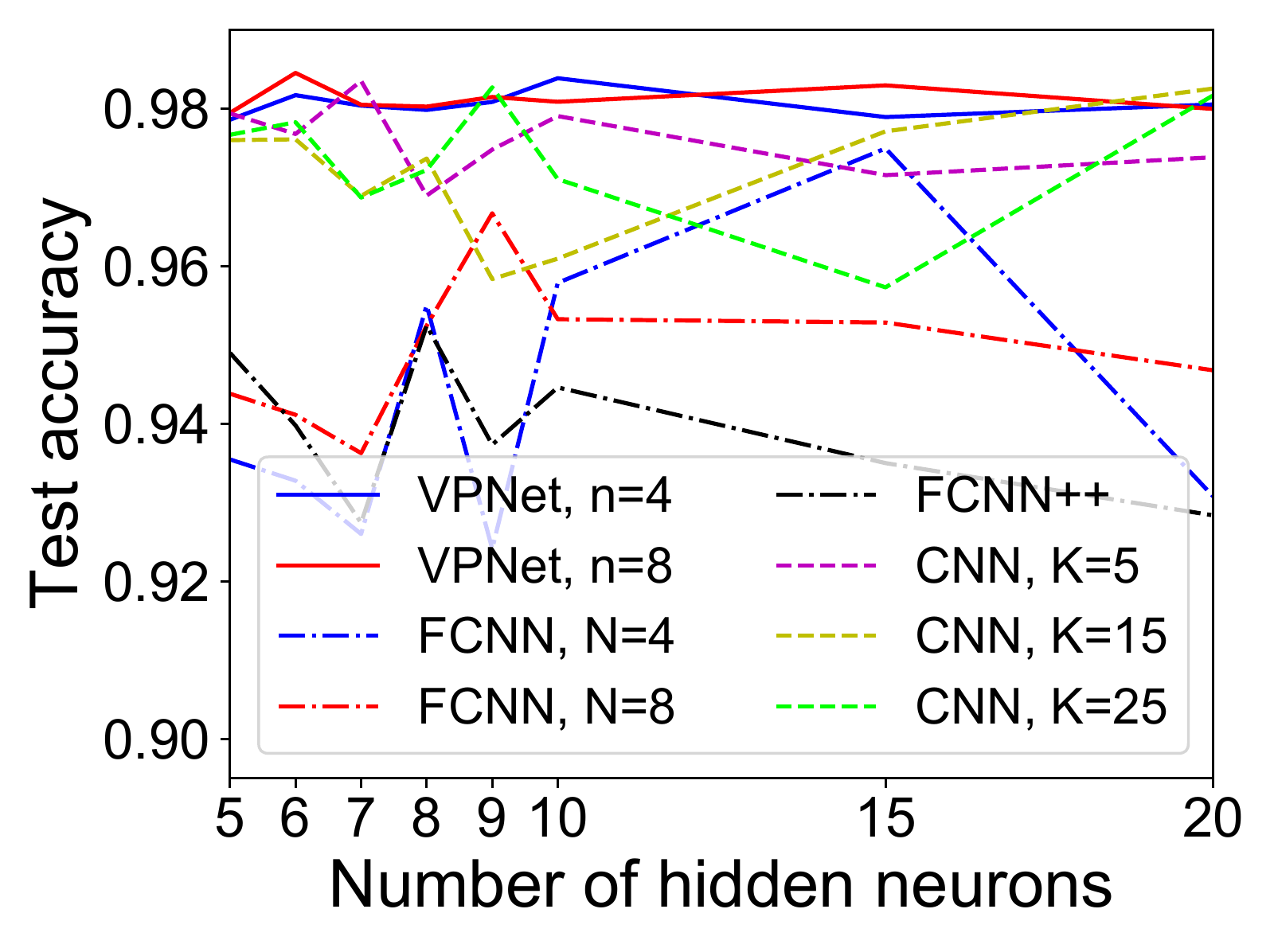}}
\hfil
\caption{Evaluation on real data, best test accuracies}
\label{fig:ecg_results}
\end{figure*}

The performace of VPNet was measured in a similar way as in the synthetic case, with more than 3500 possible hyperparameter configurations examined. The aggregated results are presented in Fig.~\ref{fig:ecg_results}~(a) and (b), for the balanced and unbalanced case, respectively. Here, the FCNN and CNN cases were restricted so that the output dimensions of the first layer were similar to the VP dimensions, and only the number of neurons in the hidden layer were varied. Note that VPNet again required a remarkably low number of network parameters. We also evaluated another, larger FCNN configuration (FCNN++), where the number of neurons in the first layer was not restricted to that of the VP dimension $n$, but had the same number as in the second, hidden layer. The structure and distribution of training and test data were more complex than in the synthetic case, which clearly made the classification task more difficult for all network architectures. Again, we conclude that VPNet can outperform FCNNs and CNNs for low-complexity networks. Note that VPNet reaches peak performance at low network complexity (at low number of hidden neurons, i.e. at low number of system parameters), and the performance starts to decrease early if we increase the complexity. This behaviour is slightly different for CNNs and FCNNs. A possible reason behind is that the first layer of the VPNet acts as a model-based feature extraction, i.e. provides a low-dimensional sparse representation of the input (4 or 8 features for 100 samples). Increasing the complexity of the fully-connected layers of VPNet without increasing the VP parameters or features will lead to over-parametrization and overfitting.

In addition to the total accuracies, the usual performance metrics are also provided in Table~\ref{tab:ecg_results}. Namely, sensitivity/precision ($Se$) and positive predictivity/recall ($+P$) was evaluated for each classes, as
\begin{equation}
Se = \frac{TP}{TP+FN} \quad\text{and}\quad +P = \frac{TP}{TP+FP},
\end{equation}
where $TP$, $FP$, and $FN$ are the true positive, false positive, and false negative matches, respectively. Reference intervals of the state-of-the-art are also given according to the survey Ref.~\cite{ECGsurvey}. We note that the direct comparison is not always possible, since most of these results refer to a 3 or 5 class classification of the database.

\section{Conclusion}
\label{sec:conc}
We developed a novel model-driven NN which incorporates expert knowledge via variable projections. The VP layer is a generic, learnable feature extractor or filtering method that can be adjusted to several 1D signal-processing problems by choosing an application-specific function system. The proposed architecture is simple, which means it has only a few, interpretable parameters. Our case studies showed that VPNet can achieve similar or slightly better classification accuracy than its fully connected and CNN counterparts while using a smaller number of parameters. In our tests, the convergence of the VPNet was slightly better than that of the CNN and the FCNN counterparts. However, the VP layer required only two parameters for learning in all cases, whereas the number of weights and biases for the FCNN and CNN grew linearly with the length of the input signals. These results show that VPNet can be applied effectively to various problems in 1D signal processing including classification, regression, and clustering, which we will investigate as part of future work.

%Finally, note that the construction of VPNet is general in nature, since prior knowledge is incorporated via the basis functions $\Phi_k$ and their parameterization $\theta$. In fact, several applications can be reformulated as a VP problem \cite{golub_pereyra2003, SNLLS2021}. For instance, the function system $\Phi_k(t;\tau_k,\lambda_k)=\cos(\lambda_k t + \tau_k)$ can be used in frequency estimation and in electroencephalography (EEG), where the network would learn dominant frequencies $\lambda_k$ and phases $\tau_k$ that characterize a certain class of signals, such as seizures in EEG recordings \cite{EEGseizureclass}. MRI imaging is another setting \cite{varpro_mri}, where $\Phi_k(t;\lambda_k)=\exp(-\lambda_k t)$ with $\lambda_k\in\IR^+$ yields information about the tissue type (see Section~\ref{sec:vpforwardprop_exp}). VPNet can be configured for a variety of problem domains by choosing an application-specific function systems and a suitable parametrization.

\section*{Broader Impact}
We have proposed a new compact and interpretable neural network architecture that can have a broader impact in mainly two fields: machine learning and signal processing. The
key idea is to create a network that combines the representation abilities of variable projections and the prediction
abilities of NNs in the form of a composite model. 
This concept can be generalized to other machine learning algorithms. For instance,  
VP-SVM, and other combined VP methods, such as VP-K-means and VP-C-means, can extend the potential areas of application, including classification, regression, and clustering problems. Since the nonlinear parameters of the VP layer are interpretable, they can also be used in feature-space augmentation, where new data is generated from existing one in order to improve the generalization properties of DNNs \cite{dataaug, feataug}.

%These VP combined techniques may have a broader impact in mainly two fields: signal processing and machine elarnng.

Signal-processing aspects of VPNet were discussed in the ECG heartbeat classification case study. Additionally, VPNet may have great potential in a wide range of applications especially where VP has proven to be an efficient estimation method (cf. Section~\ref{sec:hermite}). Note that many already existing adaptive signal models have been reformulated as VP problems \cite{golub_pereyra2003, SNLLS2021}; however, parameterized wavelets \cite{burrus} have not yet been studied in this context. Therefore, we encourage researchers to study this class of wavelets in the framework of VPNet. 

Model-driven neural network solutions can have a great impact in biomedical engineering and healthcare informatics, where medical data classification alone is usually not enough, as physiological interpretation and explainability of the results are also important. However, special care should be taken to avoid automation bias when these approaches are applied to real-world problems \cite{autbias}. These clinical decision-support systems are difficult to validate, since this requires medical expertise and vast amounts of data. The latter is naturally unbalanced in the sense that one class of signals (e.g.\ from healthy patients), is overrepresented compared to the others. In order to address these potential biases, VPNet should be tested in various scenarios that include, for instance, noisy and incomplete measurements, or unbalanced data. 

\nonumsection{Acknowledgments} \noindent This work has been supported by the “University SAL Labs” initiative of Silicon Austria Labs (SAL) and its Austrian partner universities for applied fundamental research for electronic based systems;
and the COMET-K2 "Center for Symbiotic Mechatronics" of the Linz Center of Mechatronics (LCM), funded by the Austrian federal government and the federal state of Upper Austria; and EFOP-3.6.3-VEKOP-16-2017-00001: Talent Management in Autonomous Vehicle Control Technologies – The Project is supported by the Hungarian Government and co-financed by the European Social Fund. This paper was supported by the J\'anos Bolyai Research Scholarship of the Hungarian Academy of Sciences.
\nonumsection{Data Availability} \noindent The data and code that support the findings of this study are available at Ref.~\refcite{vpnetgit}.

%\appendix{Appendices}
%\noindent Appendices should be used only when absolutely necessary. They should come immediately before References. If there is more than one appendix, number them alphabetically. Number displayed equations occurring in the appendix as (A.1), (A.2), etc.:
%\begin{equation}
%\mu(n, t) = \frac{\displaystyle\sum^\infty_{i=1} 1(d_i < t, N(d_i) = n)} {\displaystyle\int^t_{\sigma=0} 1(N(\sigma) = n)d\sigma}\,\,.\label{that} \end{equation}

\appendix{\label{appendix:a}}
%Number displayed equations occurring in the Appendix as (B.1), (B.2), etc.
%skalárszorzat integráljának közelítése összeggel, a limesz végetelenben nulla, csak egy szignifikáns részen különbözik a nullátol, ezt a részt kell sűrűn mintavételezni, 4 részes ábra

Let us consider the Hilbert space $L^2(\IR)$ endowed with the usual scalar product and norm: 
\begin{equation*}
	\left\langle f,g\right\rangle= \int_{-\infty}^{\infty} f(t) g(t) \mathrm{d}t\,,\quad \left\|f\right\|_2=\sqrt{\left\langle f,f \right\rangle}\,, 
\label{eq:dotprod}
\end{equation*}
where $f,g\in L^2(\IR)$. It is well-known that the Hermite functions $\Phi_k\ (k\in\IN)$ in Eq.~\eqref{eq:hermitefun} are pairwise orthogonal, i.e., $\delta_{kj}=\left\langle\Phi_k,\Phi_j\right\rangle$, where $\delta_{kj}$ stands for the Kronecker delta symbol.

Another useful property of the Hermite functions $\Phi_k$ is that they converge rapidly to zero as $t\rightarrow \pm \infty$. Therefore, in practice, we can assume that each $\Phi_k$ has a compact support. This can be used to satisfy the approximate orthogonality relation: 
\begin{equation*}
    \delta_{kj}=\left\langle\Phi_k,\Phi_j\right\rangle\approx\int_{a}^{b} \Phi_k(t) \Phi_j(t) \mathrm{d}t\,,
\label{eq:truncdot}
\end{equation*}
provided that the supports of both $\Phi_k$ and $\Phi_j$ are embedded in a finite interval $[a;b]$. Note that the first Hermite function $\Phi_0(t)$ is equal, up to a constant factor, to the probability density function of the standard normal distribution $\mathcal{N}(0, 1)$. Therefore, the three-sigma rule applies, which means that around $68\%$, $95\%$, $99\%$ of the overall integral of $\Phi_0(t)$
lies within the intervals $[-\ell; \ell]$ for $\ell= 1, 2, 3$, respectively. Therefore, in the case of $k=0$, we choose the sampling interval such that $[-3;3]\subseteq[a;b]$ holds. For larger indices $k>0$, a heuristic empirical relation $\text{supp}(\Phi_k)\sim 1.05^k\cdot[-3;3]\subseteq[a;b]$ can be applied. 

In practice, we typically use only the first few Hermite functions to model compactly supported waveforms (see Section~\ref{sec:experiments}), therefore the condition  $[-3;3]\subseteq[a;b]$ is sufficient. The same reasoning applies to the scalar product of the adaptive Hermite functions:
\begin{equation*}
    \left\langle\Phi_k(\cdot;\tau,\lambda),\Phi_j(\cdot;\tau,\lambda)\right\rangle\approx\int_{\lambda(a-\tau)}^{\lambda(b-\tau)} \Phi_k(s) \Phi_j(s) \mathrm{d}s,
\label{eq:truncdotadapt}
\end{equation*}
where we simplified the integral on the right hand side by substitution $s=\lambda(t-\tau)$. In order to satisfy the approximate orthogonality relation, the condition $[-3;3]\subseteq[\lambda(a-\tau);\lambda(b-\tau)]$ must hold, i.e.:
\begin{equation*}
    -3\geq \lambda(a-\tau), \quad \text{and} \quad \lambda(b-\tau)\geq 3\,,
%\label{eq:truncdotadapt}
\end{equation*}
which implies the feasible set $\Gamma$ in Section.~\ref{sec:vpforwardprop_Hermite}.

In Fig.~\ref{fig:cond_example}, we show four realizations of the first three adaptive Hermite functions sampled at $m=1000$ number of equidistant points. The top figures demonstrates ideal cases when $(\tau,\lambda)\in\Gamma$, whereas the bottom figures show extreme examples with too large translation $\tau$ and too small dilation $\lambda$.  

\begin{figurehere}
\centering
\includegraphics[scale=0.38, trim=28 200 30 150, clip]{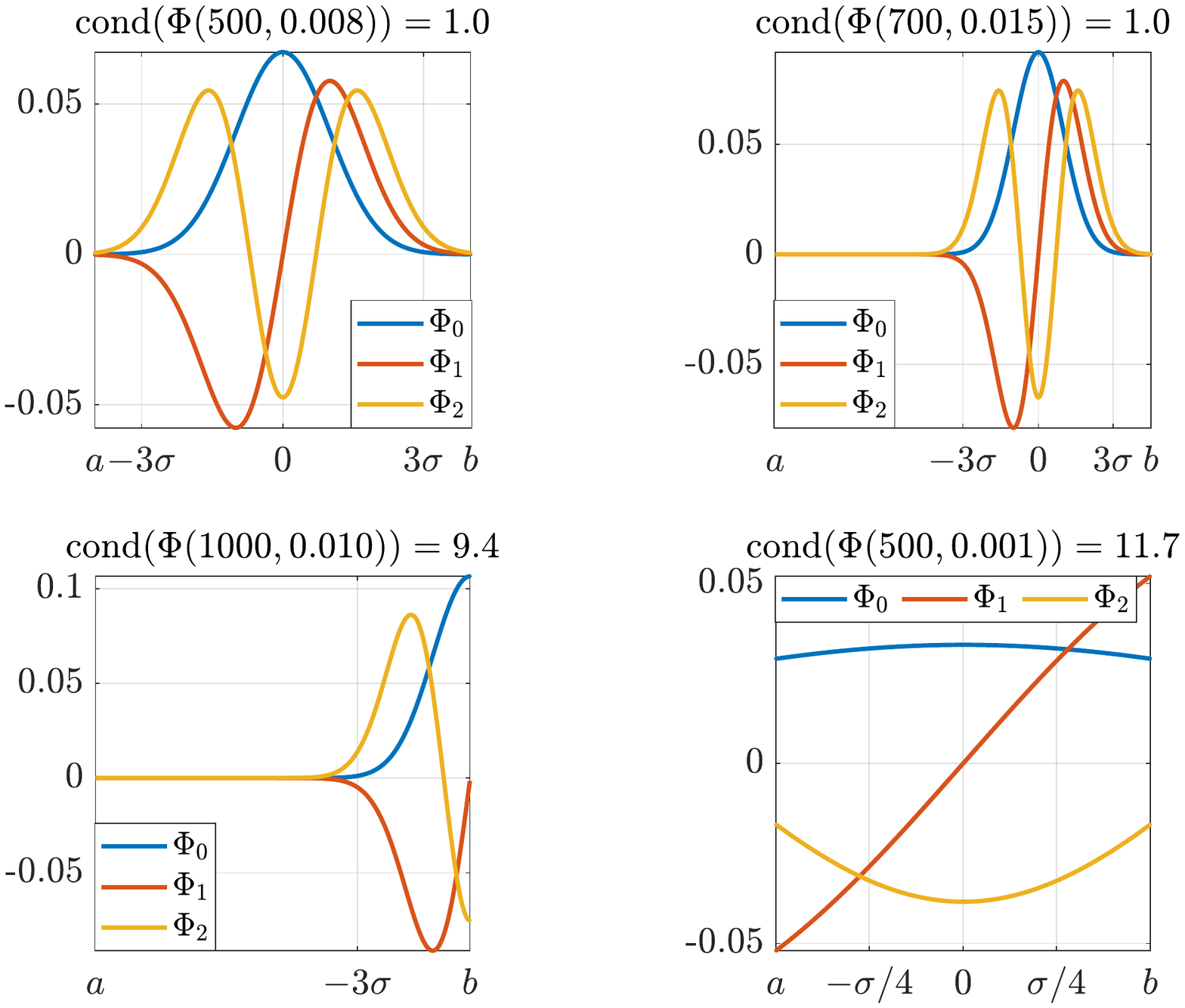}
\label{fig:cond_example}
\caption{Ideal and extreme cases for $\text{cond}(\bs{\Phi}(\tau,\lambda))$.}
\end{figurehere}

\bibliographystyle{ws-ijns}
\bibliography{references}

\end{multicols}
\end{document}